\documentclass[letterpaper]{article}
\usepackage{uai2020}
\usepackage[margin=1in]{geometry}

\makeatletter
    \let\@internalcite\cite
    \def\cite{\def\citeauthoryear##1##2{##1, ##2}\@internalcite}
    \def\shortcite{\def\citeauthoryear##1{##2}\@internalcite}
    \def\@biblabel#1{\def\citeauthoryear##1##2{##1, ##2}[#1]\hfill}
\makeatother

\usepackage{times}
\usepackage{multirow}
\usepackage{microtype}
\usepackage{graphicx}
\usepackage{booktabs} 
\usepackage{subcaption}

\usepackage{mathtools}
\usepackage{amsmath}
\newtheorem{definition}{Definition}
\newtheorem{hyp}{Hypothesis}
\usepackage{tcolorbox}
\newtcolorbox{mybox}{left=2.5pt,right=2.5pt}

\usepackage{tikz}
\usetikzlibrary{arrows.meta,calc,decorations.markings,math,positioning,decorations.pathreplacing,calligraphy}

\usepackage{hyperref}

\title{Statistical Equity: A Fairness Classification Objective}

\author{} 

\author{ {\bf Ninareh Mehrabi} \\
University of Southern California \\
Information Sciences Institute\\
ninarehm@usc.edu \\
\And
{\bf Yuzhong Huang}  \\
University of Southern California \\
Information Sciences Institute\\
yuzhongh@isi.edu
\And
{\bf Fred Morstatter}   \\
University of Southern California \\
Information Sciences Institute\\
morstatt@usc.edu
}

\begin{document}

\maketitle


\begin{abstract}
Machine learning systems have been shown to propagate the societal errors of the past. In light of this, a wealth of research focuses on designing solutions that are ``fair.'' Even with this abundance of work, there is no singular definition of fairness, mainly because fairness is subjective and context dependent. We propose a new fairness definition, motivated by the principle of equity, that considers existing biases in the data and attempts to make equitable decisions that account for these previous historical biases. We formalize our definition of fairness, and motivate it with its appropriate contexts. Next, we operationalize it for equitable classification. We perform multiple automatic and human evaluations to show the effectiveness of our definition and demonstrate its utility for aspects of fairness, such as the feedback loop.
\end{abstract}

\section{Introduction}
With the omnipresent use of machine learning in different decision and policy making environments, fairness has gained significant importance. This became the case when researchers noticed that an AI system used to measure recidivism risk in bail decisions was biased against certain racial groups \cite{angwin2016machine}. As a reaction to the disclosure of this issue and various others, the AI community has made efforts to mitigate biased and unfair outcomes in decision making processes. Many researchers have proposed definitions of algorithmic fairness, while others have tried to use these definitions in different down-stream tasks in an effort to overcome unfair outcomes. Despite the abundance of fairness definitions, the majority of them are not complete \cite{gajaneformalizing}. Moreover, theoretical analysis of these definitions have found that many at the forefront are incompatible with each other \cite{kleinberg2016inherent}. For now at least, fairness remains a philosophical question that is not yet answered in the computational domain.
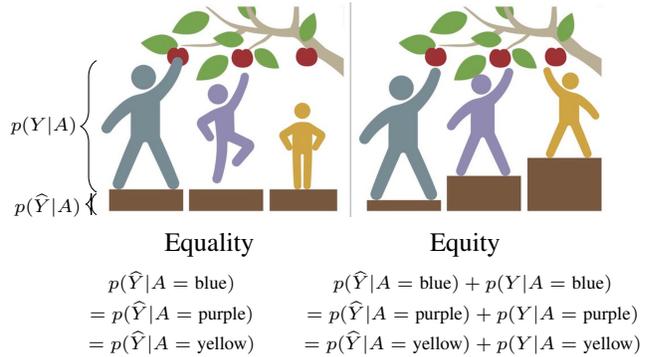
\begin{figure}[t]
\hspace{-1em}
\begin{tikzpicture}
\node[inner sep=0pt] (image) at (0,0) {\includegraphics[width=0.85\columnwidth]{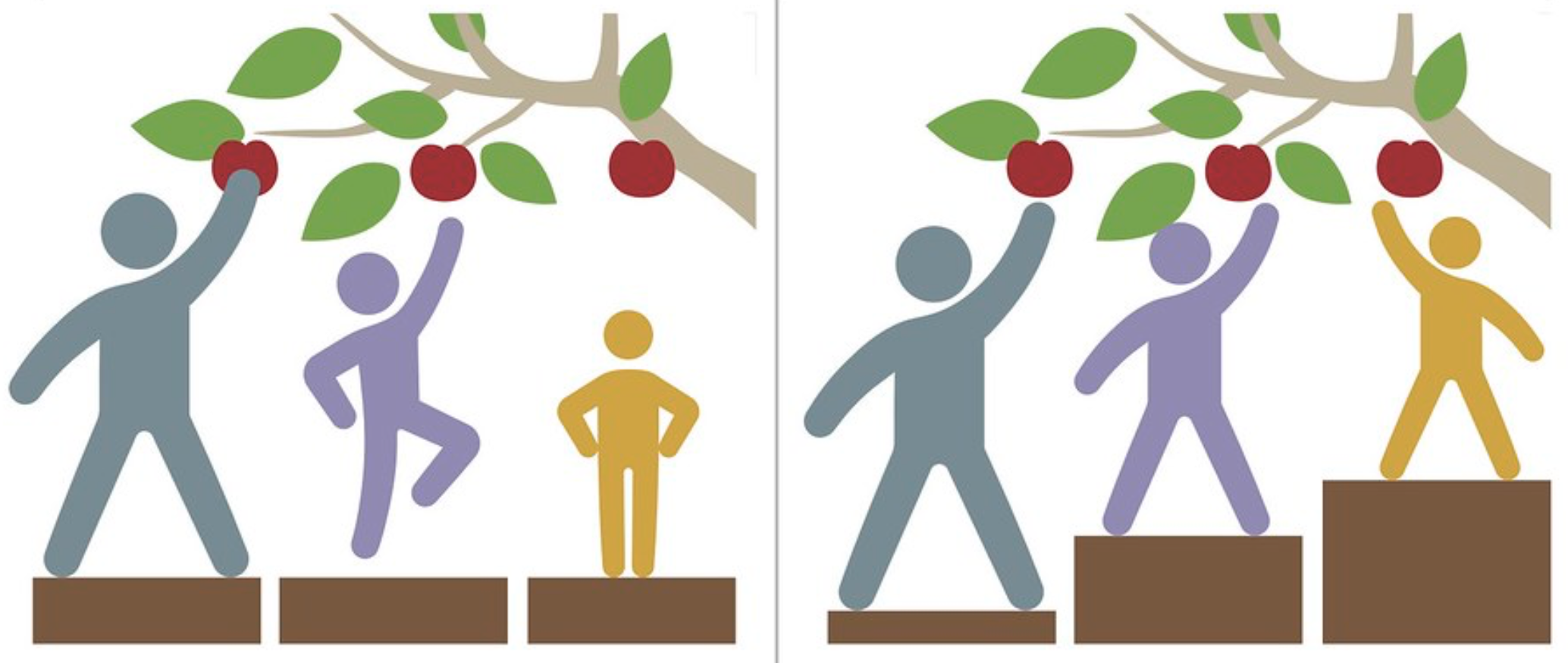}};

\draw (-1.9, -1.8) node {Equality};
\draw (+1.5, -1.8) node {Equity};

\scriptsize
\draw (-2.4, -2.3) node {$p(\widehat{Y} | A=\text{blue})$};
\draw (-2.4, -2.7) node {$=p(\widehat{Y} | A=\text{purple})$};
\draw (-2.4, -3.1) node {$=p(\widehat{Y} | A=\text{yellow})$};

\draw (+1.6, -2.3) node {$p(\widehat{Y} | A=\text{blue}) + p(Y | A=\text{blue})$};
\draw (+1.6, -2.7) node {$=p(\widehat{Y} | A=\text{purple}) + p(Y | A=\text{purple})$};
\draw (+1.6, -3.1) node {$=p(\widehat{Y} | A=\text{yellow}) + p(Y | A=\text{yellow})$};

\draw [pen colour={black!50},decorate,decoration={brace,amplitude=6pt}]
(-3.4,-1.1) -- (-3.4,+0.65) node [black] {};
\draw [pen colour={black!50},decorate,decoration={brace,amplitude=4pt}]
(-3.4,-1.4) -- (-3.4,-1.12) node [black] {};

\draw (-4.1, -0.225) node {$p(Y|A)$};
\draw (-4.05, -1.275) node {$p(\widehat{Y}|A)$};

\end{tikzpicture}
\caption{Notion of equality in fairness is depicted and formalized along with our newly formalized notion of equity.}
\label{motivation}
\end{figure}
 In light of that, we propose and mathematically formalize the equity notion of fairness in which resources and outcomes are distributed to overcome obstacles experienced by groups in order to maximize their opportunities \cite{schement2001imagining}. In this work we take the perspective that historical biases should be compensated and disadvantaged groups should be leveraged. We then introduce a data-driven classification objective function that operationalizes the notion of equity in which existing historical biases in the training data are compensated through predictions on the test data. 
 This approach will not only target fixing biases but it will also target minimizing the feedback loop phenomenon in which the biased data contaminates the decision making outcome, and it continues to stay and grow through the system. 
 
Our definition of fairness is an augmented version of statistical parity~\cite{Dwork:2012:FTA:2090236.2090255} that we adapt to measure equity. Unlike previous definitions and objective functions in which only equality is considered in present outcomes, our approach will consider the historical biases present in the data and will combine the notion of equity and affirmative action while also satisfying equality amongst groups. Our definition is a departure from differential privacy \cite{jagielski2019differentially} and fairness through unawareness \cite{10.1145/3287560.3287594} since having access to sensitive attributes is necessary in some scenarios. For instance, in cases that pertain to the medical domain, access to sensitive attributes such as gender and age are required to make a decision.

Two different fairness realizations are depicted in Figure \ref{motivation}. On the left side there is the notion of equality in which every group is given an equal amount of resources, which is too much for some members and insufficient for others. This is the problem that motivates this work: how can a classifier produce predictions that are good for the majority of a group or society? This leads us to the right picture which depicts equity where leverage is given through the model to give the groups appropriate resources to reach their goals. \\
\textbf{Note}: We use parity (statistical parity) and equality interchangeably as synonyms throughout this paper. \\ \\
The contributions of this paper are as follows:
\begin{enumerate}
    \item We define and formalize equity as a fairness definition.
    \item We demonstrate how this definition can be made actionable by proposing a loss function that combines equity with the cross-entropy loss into the classifier.
    Through experimentation, we demonstrate how our definition compares to the objective of equality.
    \item We then discuss and experimentally show the effectiveness of our definition in mitigating the feedback loop problem \cite{chouldechova2018frontiers}.
    \item Finally, we evaluate our fairness notion against the parity notion through human annotators and show how these definitions fare in different real life scenarios.
\end{enumerate}

\section{Defining Equity}
Equity is the distribution of resources among groups to overcome obstacles and to raise their opportunities for access \cite{schement2001imagining}. Thus, historically disadvantaged groups are compensated, and others get their fair share to reach their goals. To operationalize equity, we learn the existing biases from the historical data and compensate for them in the predictions generated by the model. This can be viewed similarly to affirmative action in which present decisions (algorithmic or otherwise) are made to compensate for biases of the past but in such a way that groups reach an ultimate equality. The goal is to equalize all the groups in the long run and give each group their fair share.
We modeled this phenomenon in our classification objective function which will be described in detail in this section. \\ 
Our objective function consists of two terms:
\begin{itemize}
    \item The Fairness Objective: In which the goal is to enforce equity amongst groups.
    \item The Classification Objective: In which we enforce the classification objective to achieve predictive accuracy.
\end{itemize}
Finally, we combine these two objectives and control the importance of each using a regularization parameter. 
\subsection{Operationalizing Equity}
Herein, we formalize our equity notion of fairness. Let $Y$ be a random variable denoting an outcome of a decision making process and $A$ be the sensitive variable (e.g., demographic group membership).  
Let $D$ be the set of all decisions made in the past (or historical decisions).
Let the joint distribution $p_D(Y, A)$ summarize the essential statistics of past decisions.
We are interested in the case when past decisions contain bias, i.e., $p_D(Y\mid A=a) \not\approx p_D(Y \mid A=b)$.

Let $M$ be the set of instances for which a prediction has to be done using a machine learning method.
As mentioned earlier we want the decisions of the classifier to account for and reverse the biases in the data in an equitable manner.
We formalize this in the following way.
Let a joint distribution $p_M(Y, A)$ summarize the essential statistics of the classifier on $M$.
Our goal is to generate equitable decisions for each group:
\begin{align*}
&p_D(Y=y \mid A=a) p_D(A=a) + p_M(Y = y \mid A=a) p_M(A=a)\\ 
=&p_D(Y=y \mid A=b) p_D(A=b) + p_M(Y = y \mid A=b) p_M(A=b), 
\end{align*}

for each possible outcome $y$.
We assume that we study the same number of instances from each demographic group between both the historical and outcome sets (i.e., $p_D(A = a) = p_D(A = b) = 1/2$ and $p_M(A=a)=p_M(A=b)=1/2$).
Under this assumption, our fairness criterion becomes:
\begin{align*}
&p_D(Y=y \mid A=a) + p_M(Y = y \mid A=a)\\ 
=&p_D(Y=y \mid A=b)  + p_M(Y = y \mid A=b). 
\end{align*}
One can interpret this criterion as equalizing odds of different groups for a random person drawn with equal probability from $D$ or $M$.

To comply with the notation widely used in the fairness literature,  $p_D(Y | A=a)$ can also be written as $p(Y|A=a)$ and $p_M(Y| A=a)$ as $p(\hat{Y}| A=a)$.\\
Finally, to translate our new fairness notion into widely used format one can write our objective as follows:
\begin{align*}
    p(\hat{Y}|A=a)+p(Y|A=a)=p(\hat{Y}|A=b)+p(Y|A=b).
\end{align*}
\begin{tcolorbox}
\begin{definition} [Statistical Equity]
  A predictor is statically equitable among demographic groups, $a$ and $b$, if it satisfies $p(\hat{Y}|A=a)+p(Y|A=a)=p(\hat{Y}|A=b)+p(Y|A=b)$.
\end{definition}
\end{tcolorbox}

\subsection{Classification Objective}
In order to satisfy the fairness objective 
and to be able to couple it with our classification loss, we divided our total classification objective into two terms. The first term, denoted by $F_{equity}$, attempts to satisfy the fairness objective, and the other term, denoted by $L$, attempts to satisfy the classification loss. These two losses are then controlled and coupled by a regularization term $\beta$.

The resulting fairness objective for each possible outcome $y$ is coupled in our classifier as follows:
 \begin{align*}
        &F_\text{equity}(\theta)=\\
        &\;\;\;\;\;\;\sum_{y}\left(\Big[\frac{1}{n}\sum_{i=1}^{n}p(\hat{Y_i}=y|A=a)
        +p(Y=y|A=a)\Big]\right. \\
        &\;\;\;\;\;\;\left. -\Big[\frac{1}{n}\sum_{i=1}^{n}p(\hat{Y_i}=y|A=b) 
        +p(Y=y|A=b)\Big]\right)^2.
\end{align*}

Note that there is a difference between the notation of historical and predictive (future) outcomes. By equalizing the sum of historical plus future outcomes of one group to another, we are enforcing affirmative action and try to compensate for observed historical biases in the data by correcting and adjusting the predictive outcome so that eventually all the groups reach an equilibrium in our objective function. This equilibrium can be in terms of all the groups satisfying their goals, e.g. that in Figure \ref{motivation}. 
We defined our classification objective as the cross-entropy loss $L(\theta)$.
Notice that $L(\theta)$ can be any other loss; however, cross-entropy is used in our experiments. 
Finally, by combining the fairness objective with the classification objective, we define the whole objective of our fair classification task as follows: \\
\begin{equation}
     \min_\theta \; \beta F_\text{equity}(\theta)+ (1-\beta)L(\theta).
     \label{equity_eq}
\end{equation}
$\beta$ is a hyperparameter that controls the importance of the fairness constraint over the classification objective. By making the $\beta$ value larger we are enforcing more of the fairness (equity) constraint in our objective, while smaller $\beta$ value favors the classification accuracy over the fairness constraint.

\section{Experiments and Results}
\subsection{Model and Experimental Setup}
\begin{figure*}[!bt]
\centering
\includegraphics[width=0.45\textwidth,trim=3.0cm 18.9cm 6.5cm 2.5cm,clip=true]{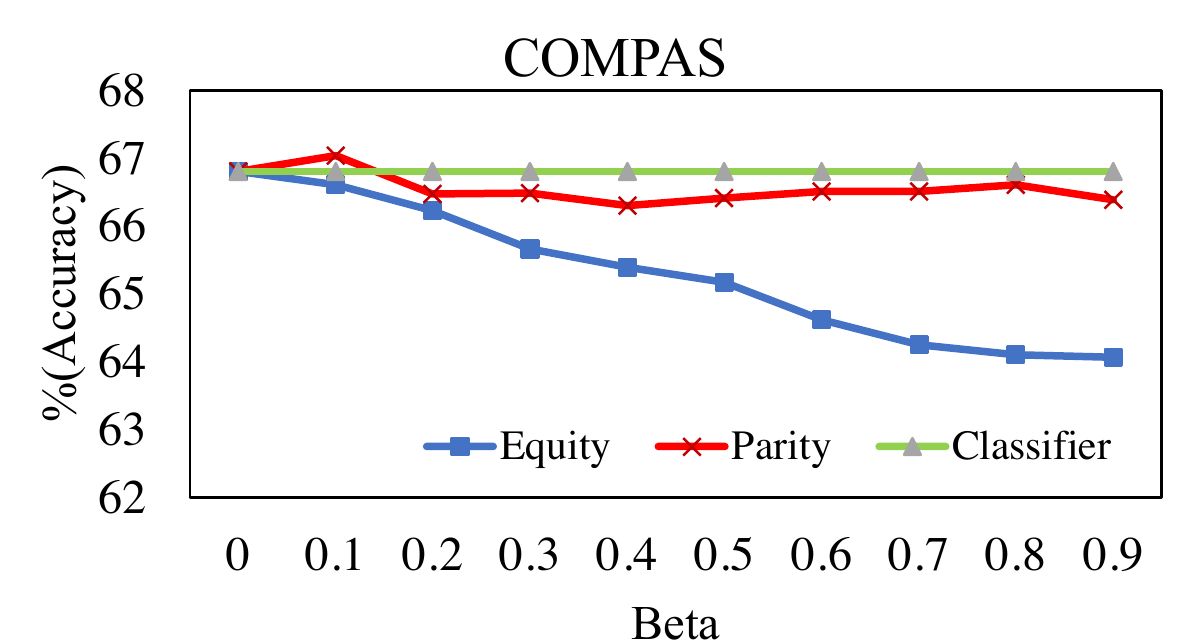}
\includegraphics[width=0.45\textwidth,trim=3.0cm 18.9cm 6.5cm 2.5cm,clip=true]{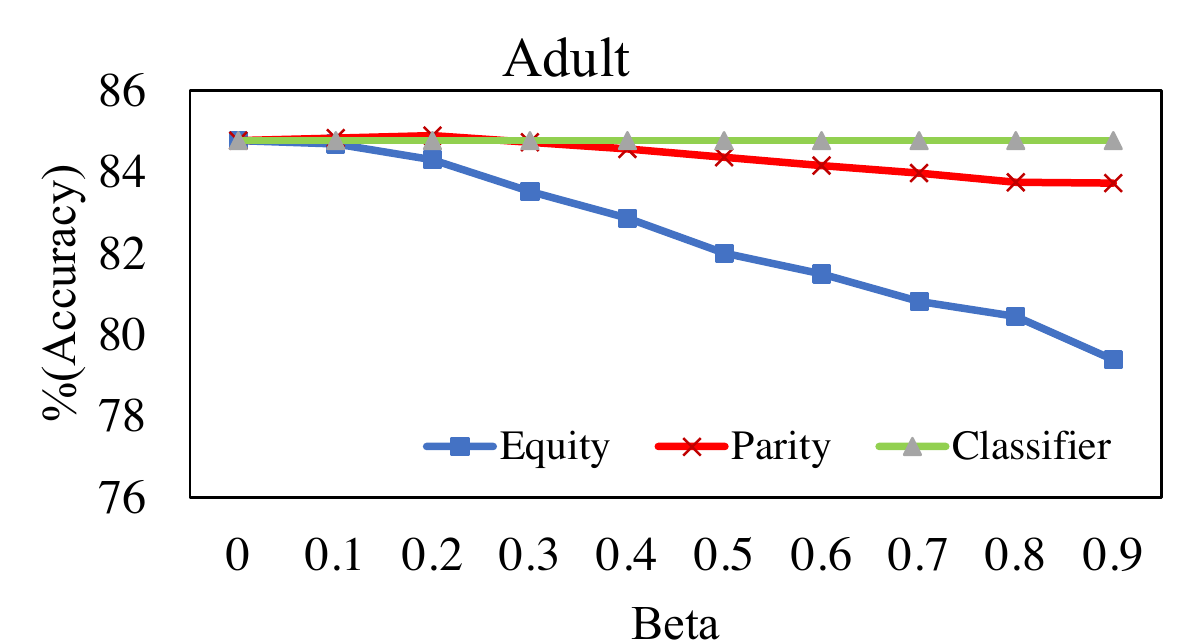}
\includegraphics[width=0.45\textwidth,trim=3.0cm 18.9cm 6.5cm 2.5cm,clip=true]{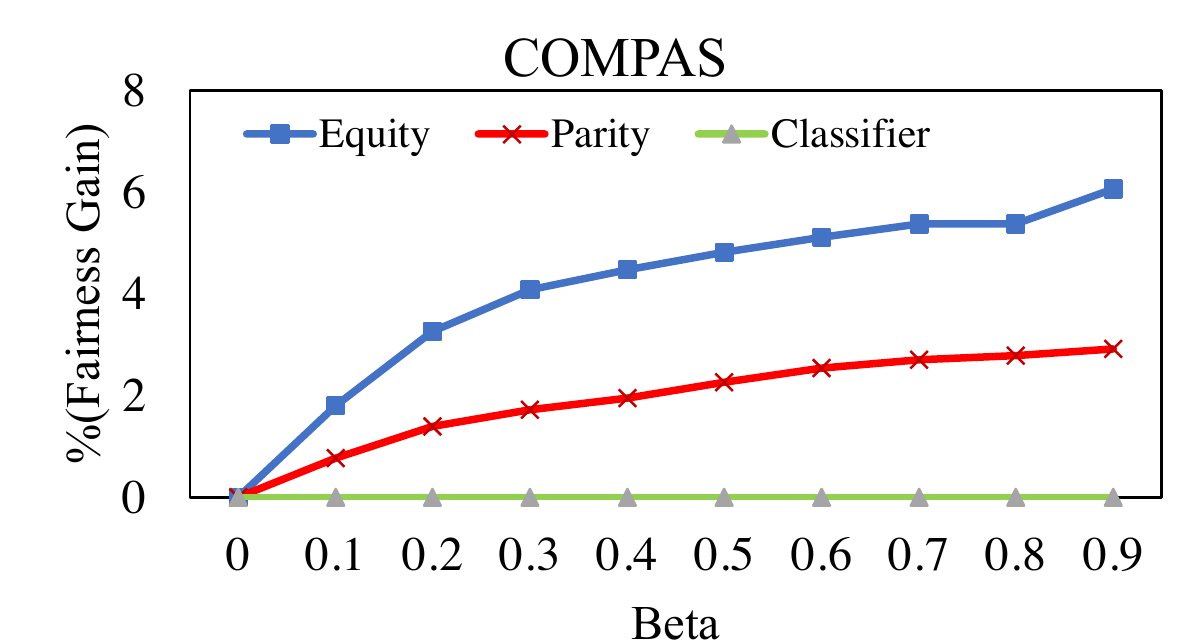}
\includegraphics[width=0.45\textwidth,trim=3.0cm 18.9cm 6.5cm 2.5cm,clip=true]{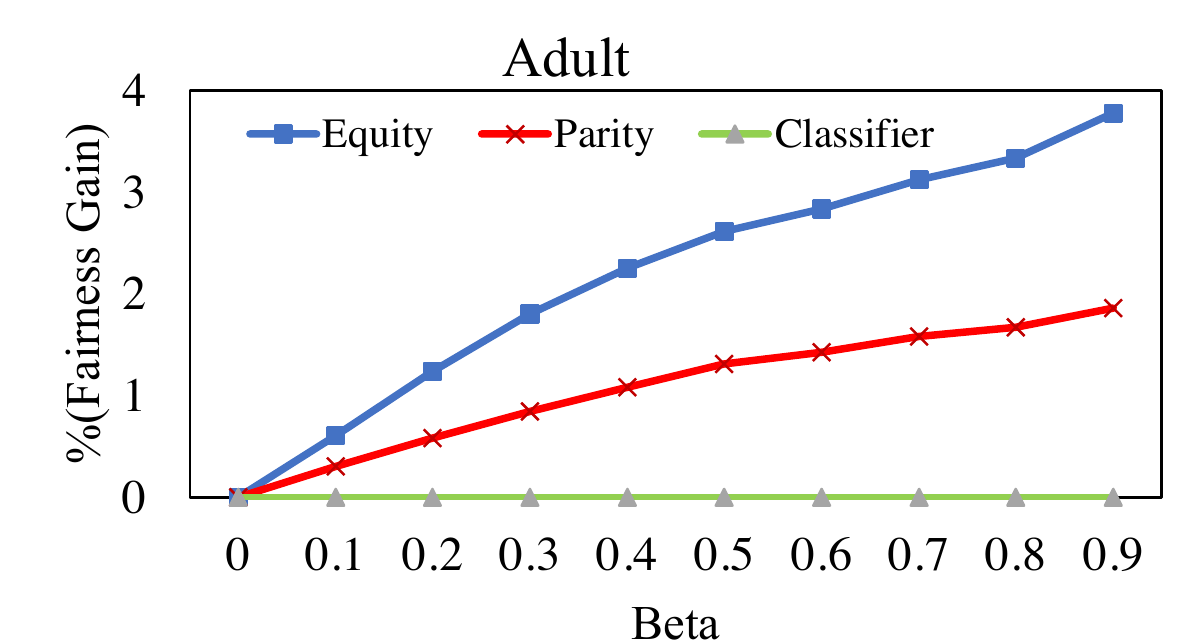}
\caption{Accuracy and fairness gain results for the COMPAS and Adult datasets over different $\beta$ values. Top plots report the accuracy results, while bottom plots report the fairness gain results. Each point on the plots is the average value of 10 experiments performed on the 10 random splits. Notice that the 10 random split sets are the same across different $\beta$ values. For details of these values along with standard deviation numbers refer to Tables \ref{Adult_acc_gain} and \ref{compas_acc_gain} in the Appendixes section.}
\label{plots1}
\end{figure*}
The model used in these experiments is a two-layer dense network with 256 hidden dimensions. We stopped training the model after seeing no improvement on the validation set for 100 iterations (validation epochs) in all the experiments with a starting learning rate of 0.01 and applied learning rate decay of 0.95. The code is available for reproduction of our experiments.\footnote{\label{projecturl}\url{https://github.com/Ninarehm/Fairness}} We performed the experiments on three different loss functions described below. 
\begin{itemize}
    \item Equity Loss: This loss function is our proposed loss function in Equation~\ref{equity_eq} introduced in the previous section.
    \item Parity Loss: This loss function is mirroring the statistical parity notion of fairness which combines the statistical parity notion with the classification loss as follows: 
    \begin{align*}
    &F_\text{parity}(\theta)=\\
    &\sum_{y}\left(\frac{1}{n}\sum_{i=1}^{n}p(\hat{Y_i}=y|A=a) \right. 
    \left.-\frac{1}{n}\sum_{i=1}^{n}p(\hat{Y_i}=y|A=b)\right)^2.
    \end{align*}

    Where $F_\text{parity}$ represents the fairness loss corresponding to the parity notion of fairness. This loss will then be combined with the classification cross-entropy loss as before which will form the whole objective loss as follows:
    \begin{equation}
     \min_\theta \; \beta F_\text{parity}(\theta)+ (1-\beta)L(\theta).
     \label{parity_eq}
    \end{equation}
    \item Classification Loss (Cross-entropy):  This is a loss function containing only the cross-entropy loss with no fairness constraints imposed as follows:
    \begin{equation}
     \min_\theta \; L(\theta).
     \label{classifier_eq}
    \end{equation}
\end{itemize}
In our results, Equity corresponds to a classifier using the equity loss function defined in Equation~\ref{equity_eq}, Parity using the parity loss function defined in Equation~\ref{parity_eq}, and Classifier using the cross-entropy loss only in Equation~\ref{classifier_eq}. We tested these classifiers on two benchmark datasets in fairness, COMPAS and Adult datasets, and reported the performance accuracy and fairness gain as defined below. \\
\begin{mybox}
\begin{definition} [Fairness Gain]
For a given loss function $\ell \in \{\text{Equity},\text{Parity},\text{Classifier}\}$, we define the fairness gain relative to a simple classifier with no fairness constraint for demographic groups $a$ and $b$ on the $D \cup M$ set as:
\begin{align*}
 \text{Fairness Gain}&=\left[|p(Y|A=a)-p(Y|A=b)|\right]_\text{classifier}\\
 & -\left[|p(Y|A=a)-p(Y|A=b)|\right]_\ell.
\end{align*}
In other words, we measure how effective a method was in reducing disparities among demographics compared to a classifier with no fairness constraint. Note how this measure is similar to the statistical parity except considering both the history and future predictive outcome.
\end{definition}
\end{mybox}
For robustness purposes, each of our experiments were performed on 10 random train, validation, and test splits for each of the datasets and the significance of our hypotheses were reported accordingly. The reported results are averages of 10 experiments performed on 10 different random splits. Due to the existing variance in the splits, we found the Mann–Whitney U tests more suitable and reliable in performing experiments. Thus, we reported the significance of the averaged results in terms of $p$-value instead of standard deviation. However, the standard deviations and detailed averaged results are all listed in the appendixes section.  
\begin{tcolorbox}
\begin{hyp} 
The Equity classification objective will achieve the highest gain in fairness, at a slight cost to accuracy. We expect this fairness gain and accuracy degrade to be more noticeable for higher $\beta$ values which controls the importance of fairness gain over classification accuracy.
\end{hyp}
\end{tcolorbox}
\subsection{COMPAS Dataset}
The COMPAS dataset contains information about defendants from Broward County. The labels in our prediction classification task were weather or not a criminal will re-offend within two years. The sensitive attribute in our experiments was gender. Among features in this dataset we used features as listed in Table \ref{compas_features}. We split the dataset into 10 different random 80-10-10 splits for train, test, and validation sets. The averaged accuracy and fairness gain results obtained from applying different losses in our classification task over 10 experiments on different splits with different $\beta$ values on the COMPAS dataset is shown in Figure \ref{plots1}.

From results shown in Figure \ref{plots1}, we can observe that classifier trained on our Equity loss is able to achieve higher fairness gain for all $\beta$ values. We also show the significance of these results in terms of one vs all (Equity vs Parity and Classifier) Mann–Whitney U test in Table \ref{fairness_p} for all the $\beta$ values. Although from the results in Figure \ref{plots1}, one can observe a degrade in performance in terms of test accuracy, the results in Table \ref{accuracy_p} show the insignificance of this degrade for low to mid $\beta$ values in this dataset. However, the degrade gets significant for higher $\beta$ values as expected due to the control of $\beta$ over accuracy-fairness trade-off. Although we reported the results for all the $\beta$ values from 0.1 to extreme of 0.9, we recommend a $\beta$ value around 0.3-0.5 which balances the fairness gain and test accuracy.
\subsection{Adult Dataset}
The Adult dataset contains information about individuals with a label corresponding if an individual's income exceeds 50k per year or not. We utilized all the features from the dataset in our classification task for predicting the label. We considered gender as the protected attribute in our classification loss. The data was split into 10 different random 80-10-10 splits for train, test, and validation sets for each set of experiments. The averaged test accuracy and fairness gain results over 10 different splits for each $\beta$ value obtained from applying different losses in our classification task on the Adult dataset is shown in Figure \ref{plots1}.

As shown in Figure \ref{plots1}, we can observe that for all $\beta$ values our definition was able to achieve higher fairness gain. We also show the significance of these results in Table \ref{fairness_p}. Although from the results in Figure \ref{plots1}, one can observe a degrade in performance in terms of test accuracy and that results in Table \ref{accuracy_p} show the significance of this degrade, this degrade is still considered to be a reasonable price for fairness considering the gain in fairness. Especially for mid $\beta$ values in which the degrade can be perceived negligible when considering the gain in fairness. As with the COMPAS dataset, we recommend a $\beta$ value around 0.3-0.5 which balances the fairness gain and test accuracy for this dataset as well. 
\begin{table}[t]
\begin{tabular}{|p{0.5cm} |c||p{1.07cm}|p{1.15cm}||p{1.07cm}|p{1.15cm}|  }
 \hline
 \multicolumn{2}{|c||}{}&\multicolumn{2}{c||}{COMPAS Dataset}&\multicolumn{2}{c|}{Adult Dataset}\\
 \hline
 \multicolumn{2}{|c||}{}&\multicolumn{2}{c||}{$p$-value}&\multicolumn{2}{c|}{$p$-value}\\
 \hline
Beta& & Parity &Classifier&Parity&Classifier\\
 \hline
0.1&\multirow{8}{*}{\rotatebox[origin=c]{90}{Equity}}   &   0.0003  &3.2e-05  &9.1e-05 &3.2e-05 \\
 0.2&    & 9.1e-05 &3.2e-05 &9.1e-05 & 3.2e-05\\
0.3&&9.1e-05 & 3.2e-05 &9.1e-05&3.2e-05\\
0.4&& 9.1e-05& 3.2e-05 &9.1e-05&3.2e-05\\
0.5&&9.1e-05 & 3.2e-05 &9.1e-05&3.2e-05\\
0.6& & 9.1e-05&3.2e-05  &9.1e-05&3.2e-05\\
0.7& &9.1e-05 &3.2e-05  &9.1e-05&3.2e-05\\
0.8&&0.0001 & 3.2e-05 &9.1e-05&3.2e-05\\
0.9&&9.1e-05 & 3.2e-05&9.1e-05&3.2e-05\\
 \hline
\end{tabular}
\caption{One vs all (Equity loss vs Parity and Classifier losses) Mann–Whitney U test for COMPAS and Adult datasets. The results show the statistical significance of experiments performed for evaluation of fairness gain amongst different losses over different $\beta$ values. The assumed test hypothesis was whether Equity will have greater fairness gain compared to Parity and Classifier losses.}
\label{fairness_p}
\end{table}

\begin{table}[h]
\begin{tabular}{ p{2.3cm} p{2.3cm} p{2cm}}
 \toprule
 Features&&\\
 \midrule
 sex&age\_cat&race  \\
 juv\_fel\_count&juv\_misd\_count&juv\_other\_count\\
 priors\_count&c\_charge\_degree&\\[0.5pt]
  \bottomrule
\end{tabular}
\caption{Features used in the experiments from the COMPAS dataset.}
    \label{compas_features}
\end{table}

\subsection{Overall Results Discussion}
As expected in our initial hypothesis, through experimentation and hypothesis testing, we were able to gain knowledge that using the Equity loss in classification will result in gain in fairness. Through Mann–Whitney U significance test we show that this gain is significant for all the $\beta$ values for both of the datasets. With regards to degrade in test accuracy, as expected, larger $\beta$ values resulted in more loss in test accuracy, while more gain in fairness. However, this loss was shown to be non-significant for one of our datasets, the COMPAS dataset, for low to mid $\beta$ values which we recommend using. For the Adult dataset, although the loss was shown to be statistically significant, the test accuracy loss was reasonable considering the price of fairness we get through the gain in fairness. Figure \ref{plots1}, demonstrates the behavior of different losses over different $\beta$ values in terms of test accuracy and fairness gain for the COMPAS and Adult datasets. Tables \ref{fairness_p} and \ref{accuracy_p} indicate the significance of our hypothesis in terms of Equity loss being able to gain highest gain in fairness and also the significance of its degrade in performance in terms of test accuracy over other baselines for the COMPAS and Adult datasets respectively. From the overall results, we suggest use of $\beta$ values between 0.3-0.5 when using our Equity objective as they are shown to be the most effective in terms of gain in fairness and maintaining a reasonable test accuracy. 

\begin{table}[t]
\begin{tabular}{|p{0.5cm} |c||p{1.07cm}|p{1.15cm}||p{1.07cm}|p{1.15cm}|  }
 \hline
 \multicolumn{2}{|c||}{}&\multicolumn{2}{c||}{COMPAS Dataset}&\multicolumn{2}{c|}{Adult Dataset}\\
 \hline
 \multicolumn{2}{|c||}{}&\multicolumn{2}{c||}{$p$-value}&\multicolumn{2}{c|}{$p$-value}\\
 \hline
Beta& & Parity &Classifier&Parity&Classifier\\
 \hline
0.1&\multirow{8}{*}{\rotatebox[origin=c]{90}{Equity}}   &  0.2725   &0.5151  &0.2245 &0.2476 \\
 0.2&   & 0.3383  & 0.1717& 0.0045& 0.0056\\
0.3&&0.1821 &0.0928  &0.0005&0.0003\\
0.4&& 0.1622& 0.0991 &9.1e-05&9.0e-05\\
0.5&& 0.0557& 0.0604 &9.1e-05&9.0e-05\\
0.6& &0.0203 & 0.0377 &9.1e-05&9.1e-05\\
0.7& &0.0187 & 0.0225 &9.1e-05&9.0e-05\\
0.8&&0.0020 & 0.0116 &9.1e-05&9.1e-05\\
0.9&&0.0008 & 0.0069 &9.1e-05&9.1e-05\\
 \hline
\end{tabular}
\caption{One vs all (Equity loss vs Parity and Classifier losses) Mann–Whitney U test for COMPAS and Adult datasets. The results test the statistical significance of experiments performed for evaluation of test accuracy amongst different losses over different $\beta$ values. The test reports the significance of degrade in performance of Equity loss over the other two losses in terms of test accuracy.}
\label{accuracy_p}
\end{table}

\section{Effect of Equity on Feedback Loop}
\begin{figure*}[h]
\centering
\begin{subfigure}[b]{0.33\textwidth}
\caption{$\beta=0.1$}
    \includegraphics[width=\textwidth,trim=2.7cm 18cm 6.5cm 2.8cm,clip=true]{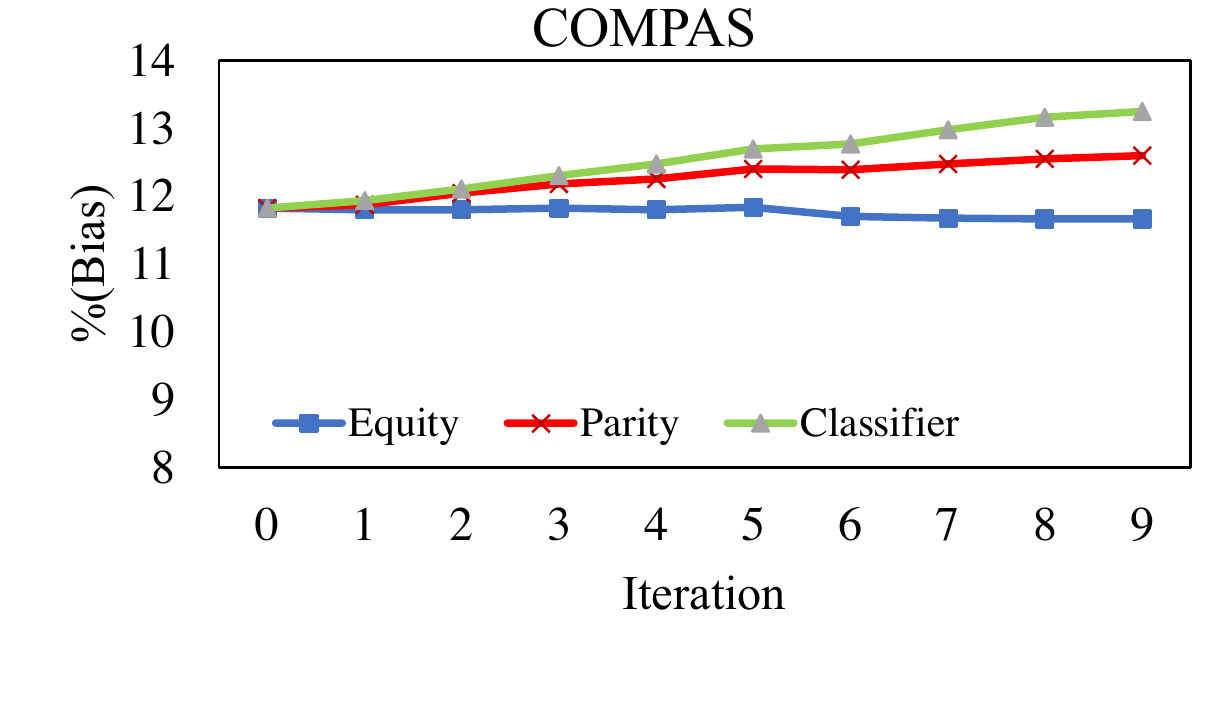}
    \end{subfigure}
    \begin{subfigure}[b]{0.33\textwidth}
\caption{$\beta=0.5$}
     \includegraphics[width=\textwidth,trim=2.7cm 18cm 6.5cm 2.8cm,clip=true]{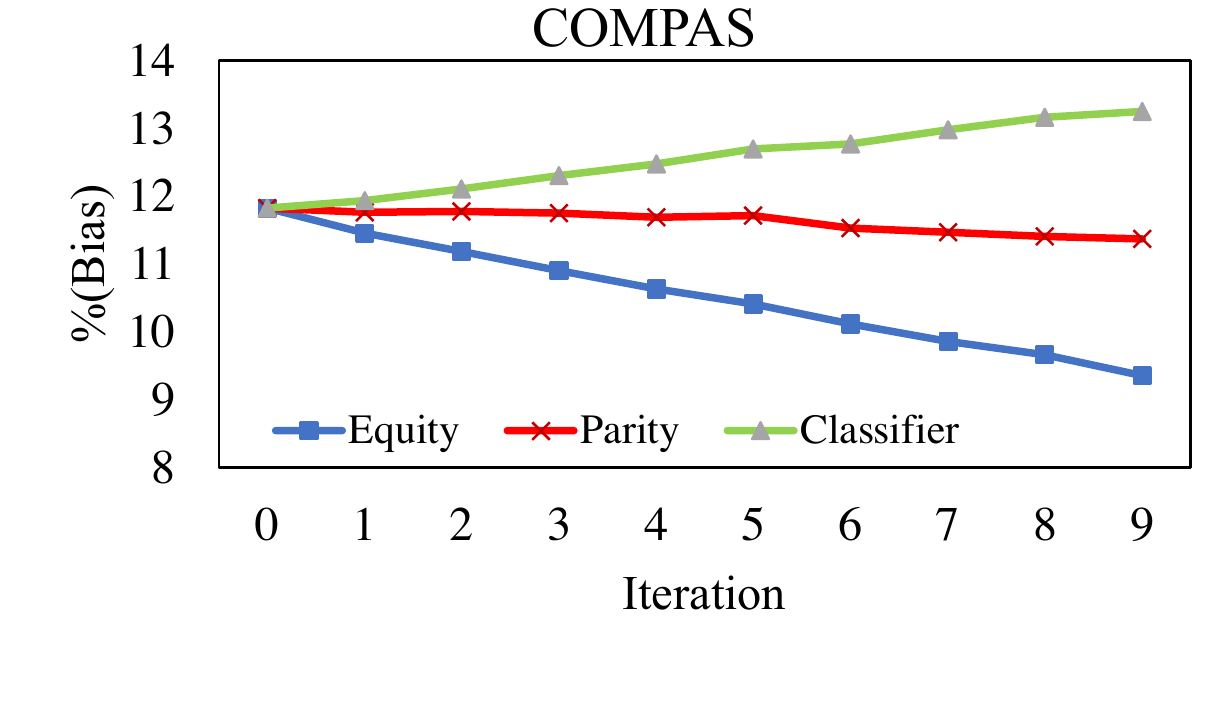}
      \end{subfigure}
     \begin{subfigure}[b]{0.33\textwidth}
\caption{$\beta=0.9$}
      \includegraphics[width=\textwidth,trim=2.7cm 18cm 6.5cm 2.8cm,clip=true]{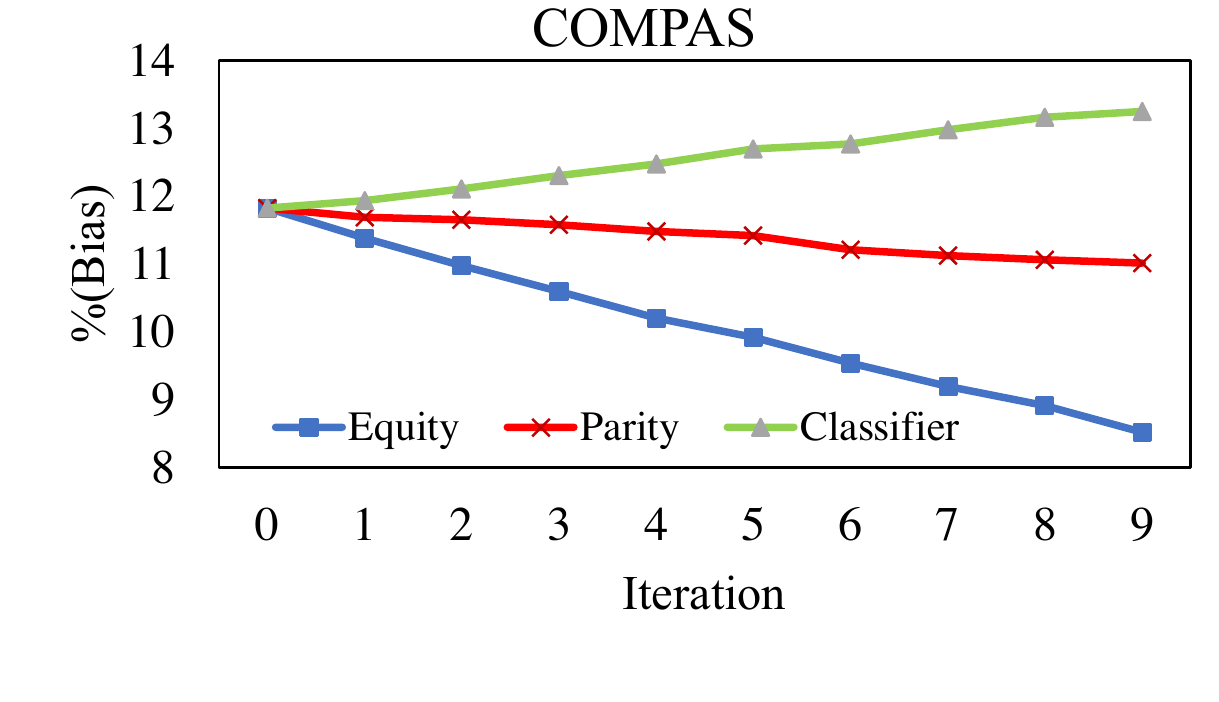}
      \end{subfigure}
      \begin{subfigure}[b]{0.33\textwidth}
      \includegraphics[width=\textwidth,trim=2.7cm 18cm 6.5cm 2.8cm,clip=true]{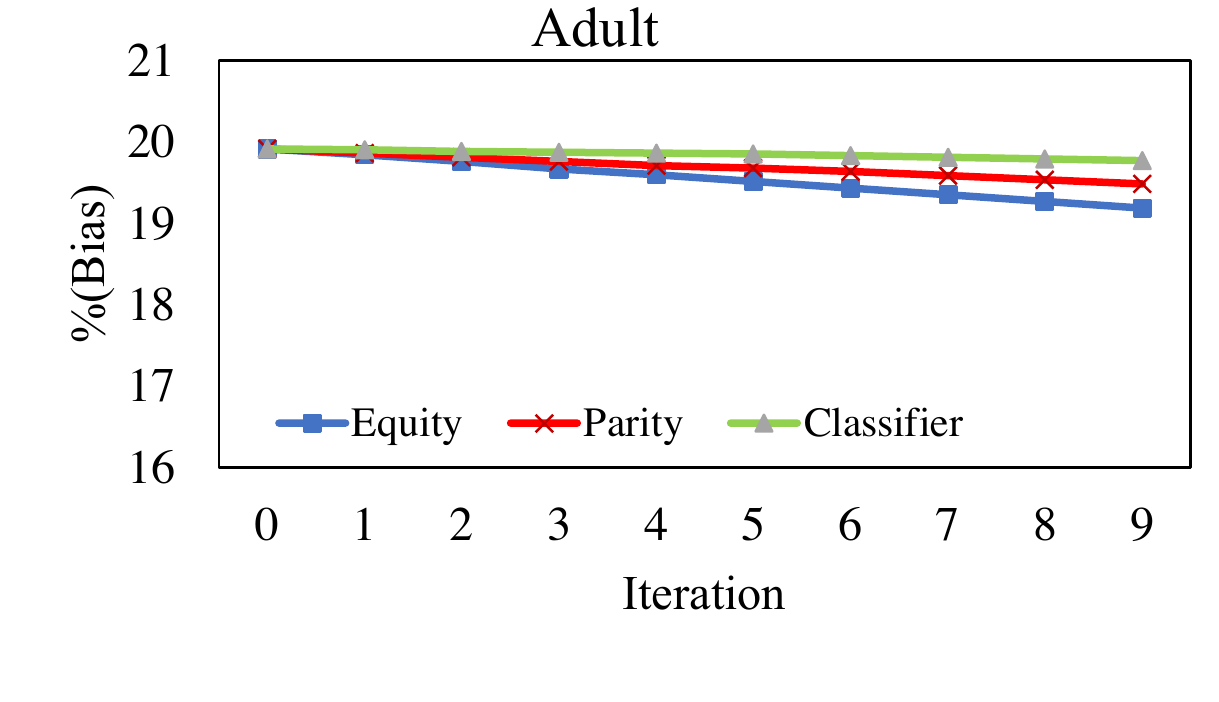}
      \end{subfigure}
      \begin{subfigure}[b]{0.33\textwidth}
        \includegraphics[width=\textwidth,trim=2.7cm 18cm 6.5cm 2.8cm,clip=true]{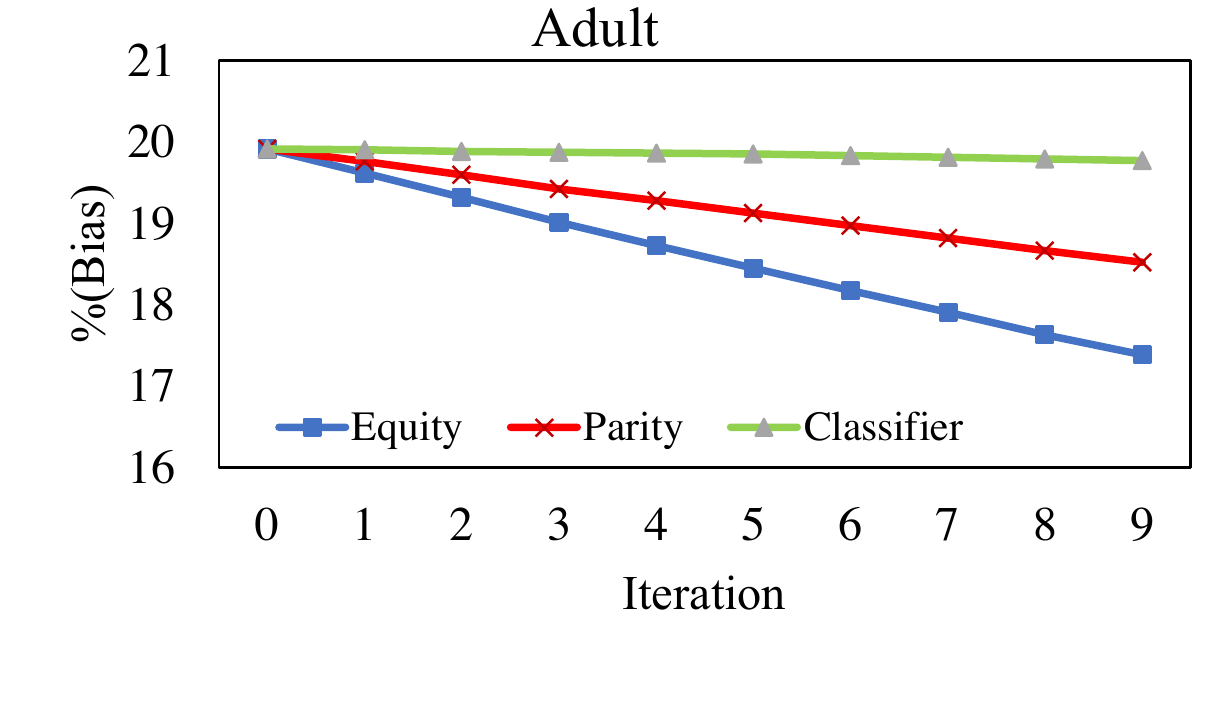}
         \end{subfigure}
        \begin{subfigure}[b]{0.33\textwidth}
         \includegraphics[width=\textwidth,trim=2.7cm 18cm 6.5cm 2.8cm,clip=true]{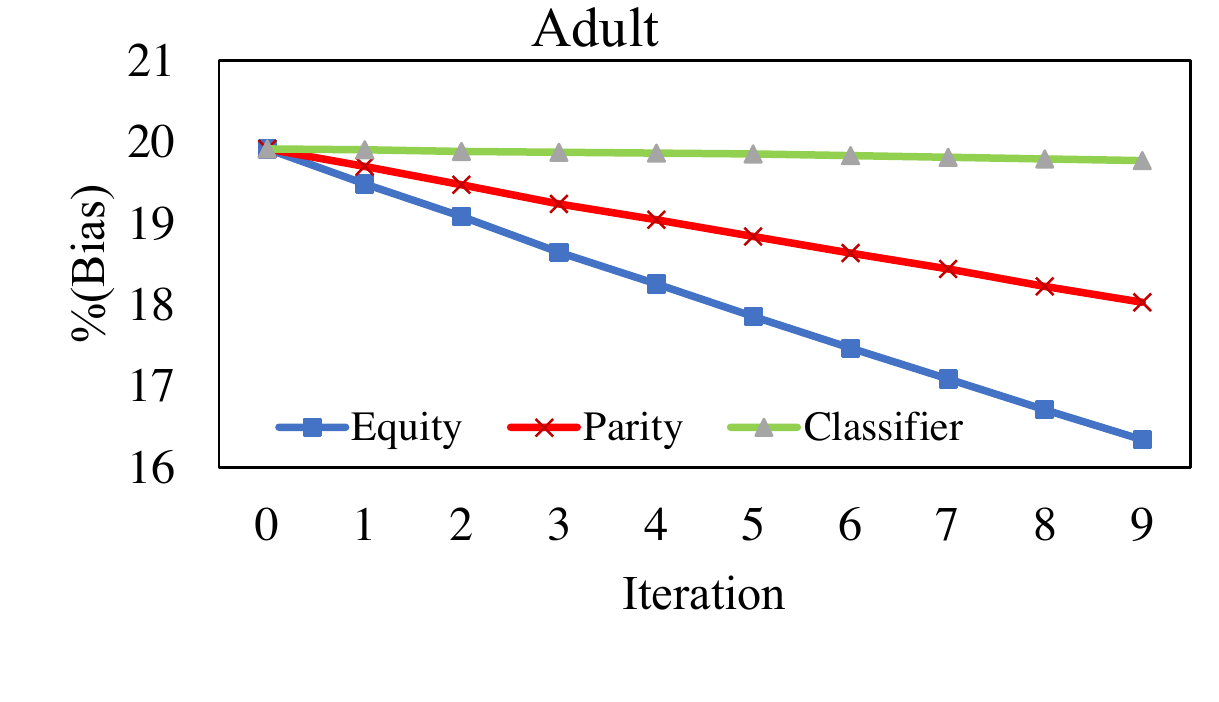}
          \end{subfigure}
    \caption{Simulation of the feedback loop phenomenon and results obtained in reduction of bias via different methods in COMPAS and Adult datasets. As expected higher $\beta$ values result in reduction of more bias in the two fairness based objectives (Equity and Parity). It also shows how Equity is more effective in reducing the bias over iterations. Each point on the plots is the average value of 10 experiments performed on the 10 random splits. Notice that the 10 random split sets are the same across different $\beta$ values.}  
    \label{iteration_exp}
\end{figure*}
An important and major concern in the fairness community is the feedback loop phenomenon \cite{chouldechova2018frontiers}. Since biased data is generated by humans, these biases are perpetuated after the models make biased decisions based on the historical biased data. The bias originates from humans, the models amplify these biases, and they loop back biased results back to the humans. This loop gets repeated and continues to carry the initial existing biases. This phenomenon is called the feedback loop phenomenon.

We hope that since our notion considers and compensates the historical biases in the training set, which might have come from humans in initial phases, and attempts to fix them by achieving an ultimate equilibrium considering the past and future decisions, it may help with the mitigation of the feedback loop phenomenon. 

In order to observe the effect of our new equity notion on fixing the historical biases in the training sets and effectively fixing the feedback loop as a consequence, we conducted experiments on datasets used in the previous section and recorded averaged results over the 10 experiments on random splits along with their Mann–Whitney U significance tests in this section with the following hypothesis. The model architecture remains the same as experiments conducted in the previous section. In addition the experiments are performed on the Equity, Parity, and Classification losses for comparison purposes.
\begin{tcolorbox}
\begin{hyp} 
The Equity classification objective can be the most effective in terms of reducing the disparities (bias) defined as $|p(Y|A=\text{a})-p(Y|A=\text{b})|$ between demographic groups $a$ and $b$ over some iterations when predictive outcomes on the test sets are accumulated over time on the historical train sets.
\end{hyp}
\end{tcolorbox}
\subsection{Experimental Design and Results}
Herein, we answer the question of what will happen if the equity classifier is allowed to play out in a realistic environment. We simulate the feedback loop as an iterative training-predicting cycle. We train our model in sequential chunks, splitting the test data into 10 equal-sized chunks. At the first iteration, we train the model using the train data. 
At each subsequent iteration, we take one of the chunks from our test data adding it to the previous train data alongside its \emph{predicted} labels and re-train the model for the next iteration. We then deleted this chunk from the test set and keep it in the train set. Each experiment was repeated 10 times with different random splits. 
\begin{table}[t]
\begin{tabular}{|p{0.5cm} |c||p{1.07cm}|p{1.15cm}||p{1.07cm}|p{1.15cm}|  }
 \hline
 \multicolumn{2}{|c||}{}&\multicolumn{2}{c||}{COMPAS Dataset}&\multicolumn{2}{c|}{Adult Dataset}\\
 \hline
 \multicolumn{2}{|c||}{}&\multicolumn{2}{c||}{$p$-value}&\multicolumn{2}{c|}{$p$-value}\\
 \hline
Iter& & Parity &Classifier&Parity&Classifier\\
 \hline
1&\multirow{8}{*}{\rotatebox[origin=c]{90}{Equity}}   & 0.1365    & 0.0809 & 0.0520& 0.0018\\
 2&    & 0.0520 &0.0106 &0.0045 &9.1e-05 \\
3&&0.0156 & 0.0004 &0.0002&9.1e-05\\
4&& 0.0070&  0.0002&9.1e-05&9.1e-05\\
5&& 0.0018& 0.0001 &9.1e-05&9.1e-05\\
6& &0.0006 & 9.1e-05 &9.1e-05&9.1e-05\\
7& &0.0008 & 9.1e-05 &9.1e-05&9.1e-05\\
8&&0.0004 &  9.1e-05&9.1e-05&9.1e-05\\
9&&0.0002 & 9.1e-05 &9.1e-05&9.1e-05\\
 \hline
\end{tabular}
\caption{Performance of Mann-Whitney U test for showing the effectiveness of Equity in reducing bias in the feedback loop compared to Parity and Classifier losses over different iterations for COMPAS and Adult datasets. The obtained $p$-values show the significance of our reported results in Figure \ref{iteration_exp} for $\beta$ value of 0.5.}
\label{05_feedbackloop_p}
\end{table}

Figure \ref{iteration_exp} reports $|p(Y|A=\text{female})-p(Y|A=\text{male})|$, averaged across 10 runs, as a measure for disparity for both predicted class labels $Y=0$ and $Y=1$ in each of the datasets for each of the fairness notions for each $\beta$ value. These results demonstrate that our notion of fairness was able to minimize the gap between $p(Y|A=\text{female})$ and $p(Y|A=\text{male})$ in all of the datasets. The results show that using our notion can bring equality, equity, and fairness in long run and mitigate the negative effects of the feedback loop phenomenon. As expected and shown in Figure \ref{iteration_exp}, higher $\beta$ values resulted in achieving more fair outcomes which resulted in reduction of bias. In addition, we reported the Mann–Whitney U test results to show the significance of our results. Table \ref{05_feedbackloop_p} shows the significance of these results for COMPAS and Adult datasets for $\beta$ value of 0.5 for different iterations supporting our hypothesis.\footnote{Results for other $\beta$ values  can be found in the \\ supplementary material.} This is consistent with our earlier finding that $\beta = 0.5$ is the most effective and reasonable with significant impact in gaining fairness, reducing bias, and balancing the fairness-accuracy trade-off.

\section{Public Perception of Equity}
In order to understand the public's perception of equity (via our proposed definition) and its comparison to equality in different real life scenarios, we conducted surveys on Amazon Mechanical Turk in the vein of \cite{saxena2019fairness}.

\subsection{Experimental Setup}
We recruited 150 workers by showing them four different real life scenarios. In each scenario, we proposed two fairness solutions: one based on equity and one based on equality/parity. For each scenario, we asked workers to rate how fair they think each solution is on a scale of zero to four. At the end of each scenario, we asked workers to select their preferred fairness solution for each scenario. We asked workers to provide written justification for their responses. In addition, we had a ``sanity check'' question at the end of our survey to discover and remove workers behaving randomly. The screenshot from our questionnaire is included in the Appendixes section for more detailed information.


A summary of the scenarios are as follows. Note that the experimental results follow the same numbering convention as listed below.
\begin{itemize}
    \item Scenario 1 (Equality vs Equity): We asked workers to rate pictures of equity and equality in Figure~\ref{motivation} and chose their preferred picture.
    \item Scenario 2 (School Loan): Workers rate loan distribution mechanisms. One is based on equity, which considers each student's past history of receiving a scholarship (equity). Another simply proposes to equally distribute the loan among all the students (parity).
    \item Scenario 3 (Government Subsidized Housing):  We asked respondents to rate the government subsidized housing distribution systems proposed in the survey---one based on equity considering how houses were historically distributed across different races (equity). The other proposes to equally distribute houses across different racial categories (parity).
    \item Scenario 4 (College Admission): We asked respondents to rate college admission systems---one based on equity considering if the student is a first generation college student (equity). The other equally admits students from first generation and non-first generation backgrounds (parity).
\end{itemize}
\subsection{Results}
After gathering and analyzing responses from mechanical turk workers, we observed that there are some cases in which our notion of fairness is strongly preferred by a large margin, and some other cases where preference is given to the parity notion. Fairness is subjective and different people may have different takes on what would be a fair solution to a particular case. That is the main reason why we introduce this notion as not only in some scenarios our definition will be over-preferred but also in some non-preferred scenarios it will get some preference from certain groups of people.

\begin{figure}
\centering
    \includegraphics[width=0.23\textwidth]{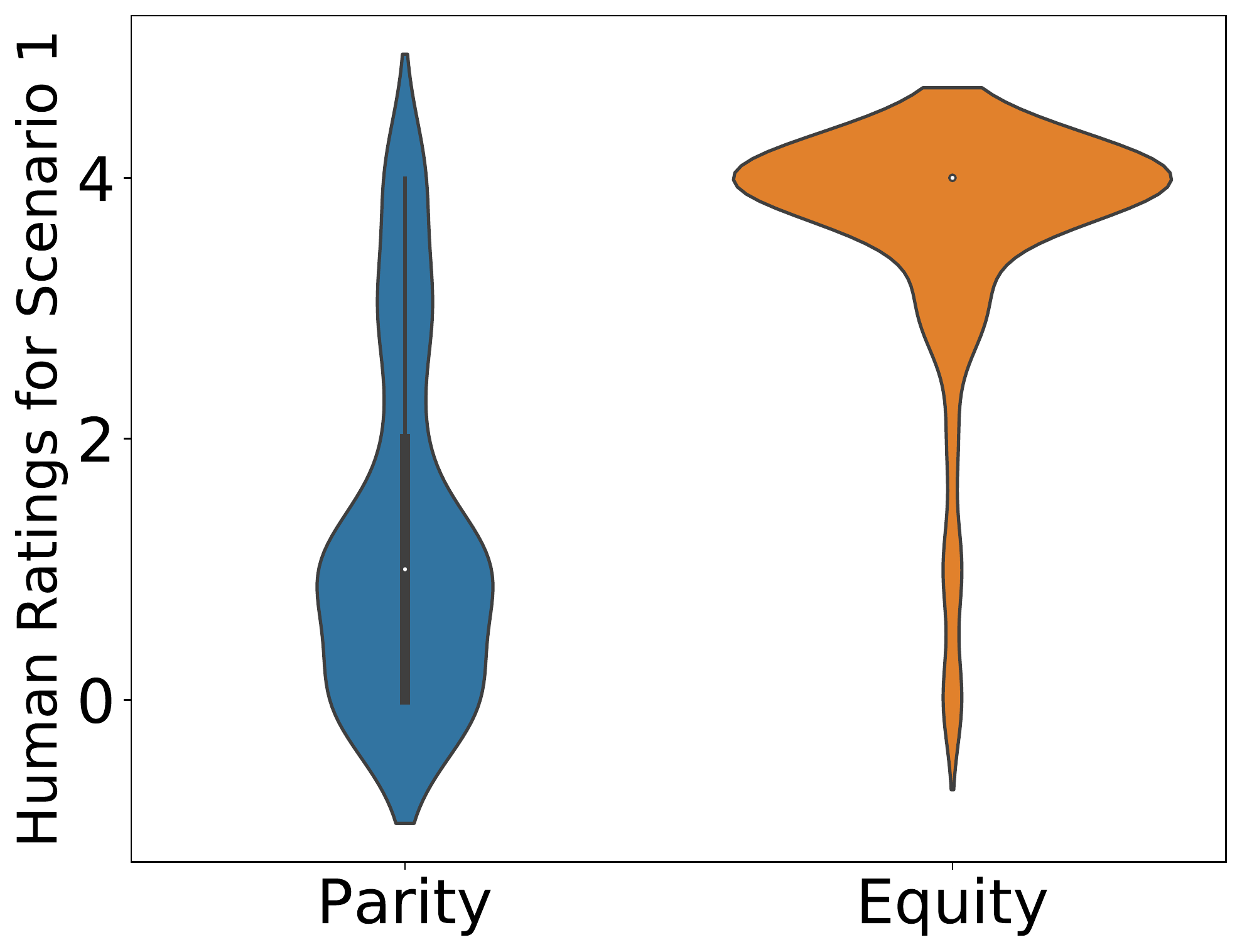}
     \includegraphics[width=0.23\textwidth]{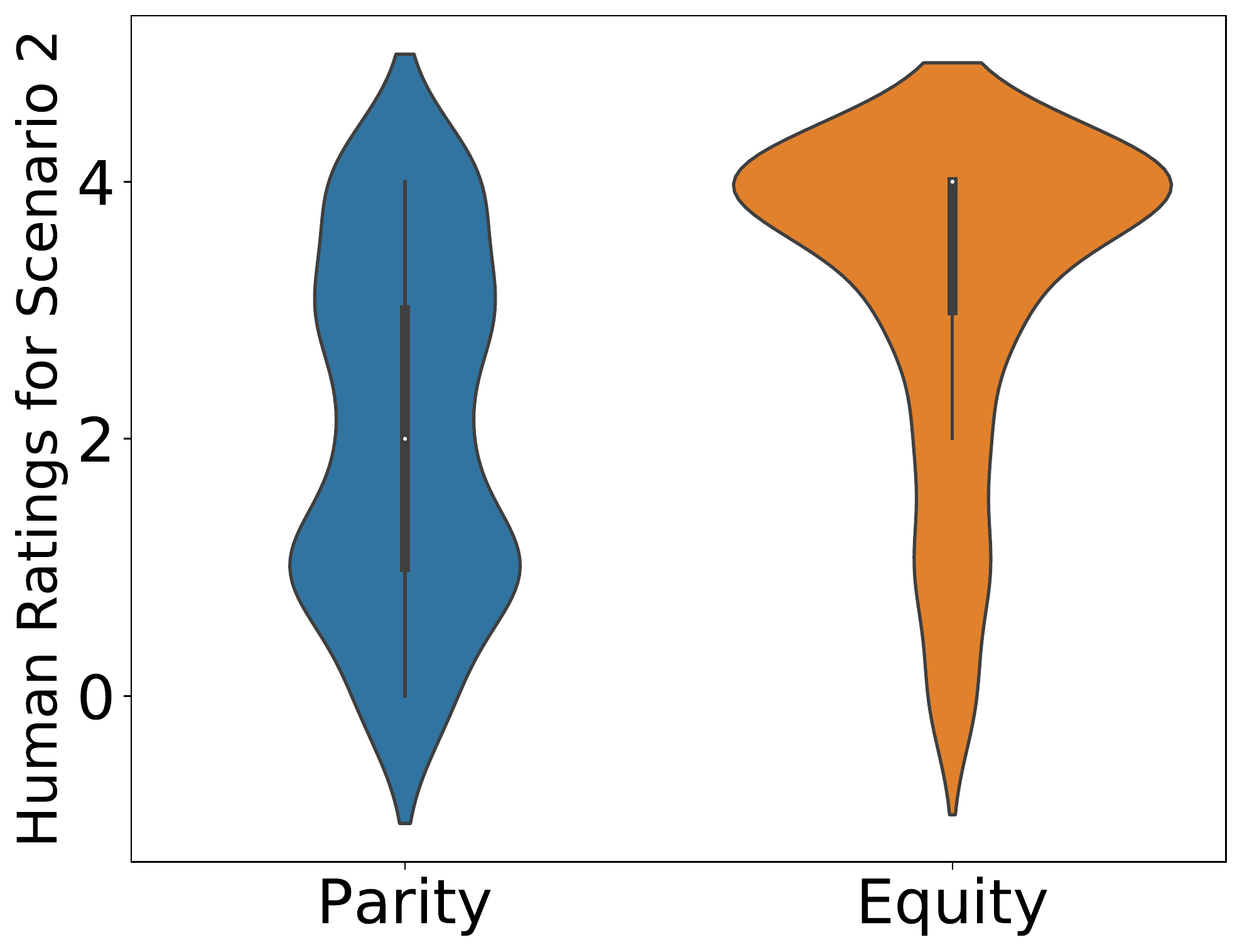}
      \includegraphics[width=0.23\textwidth]{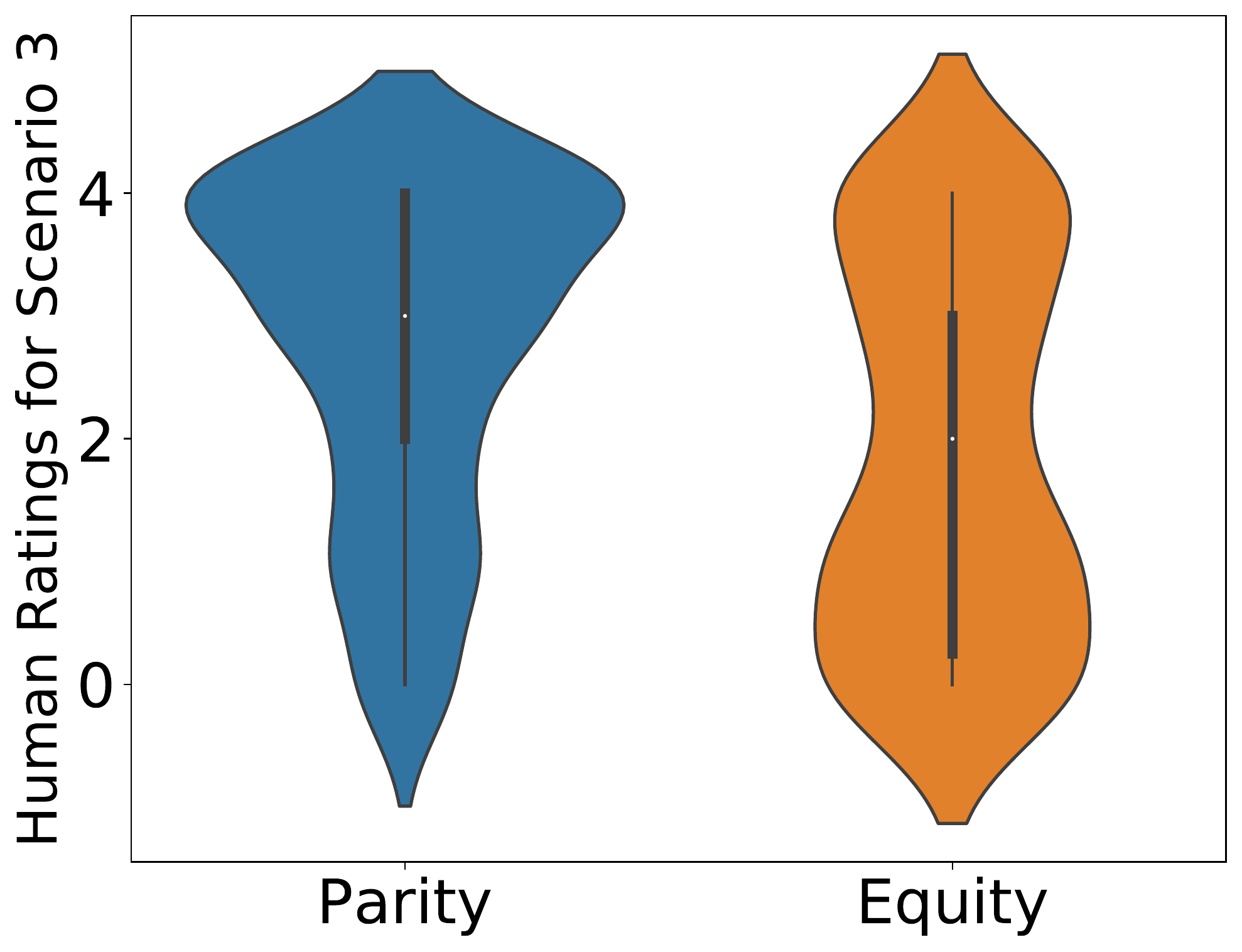}
       \includegraphics[width=0.23\textwidth]{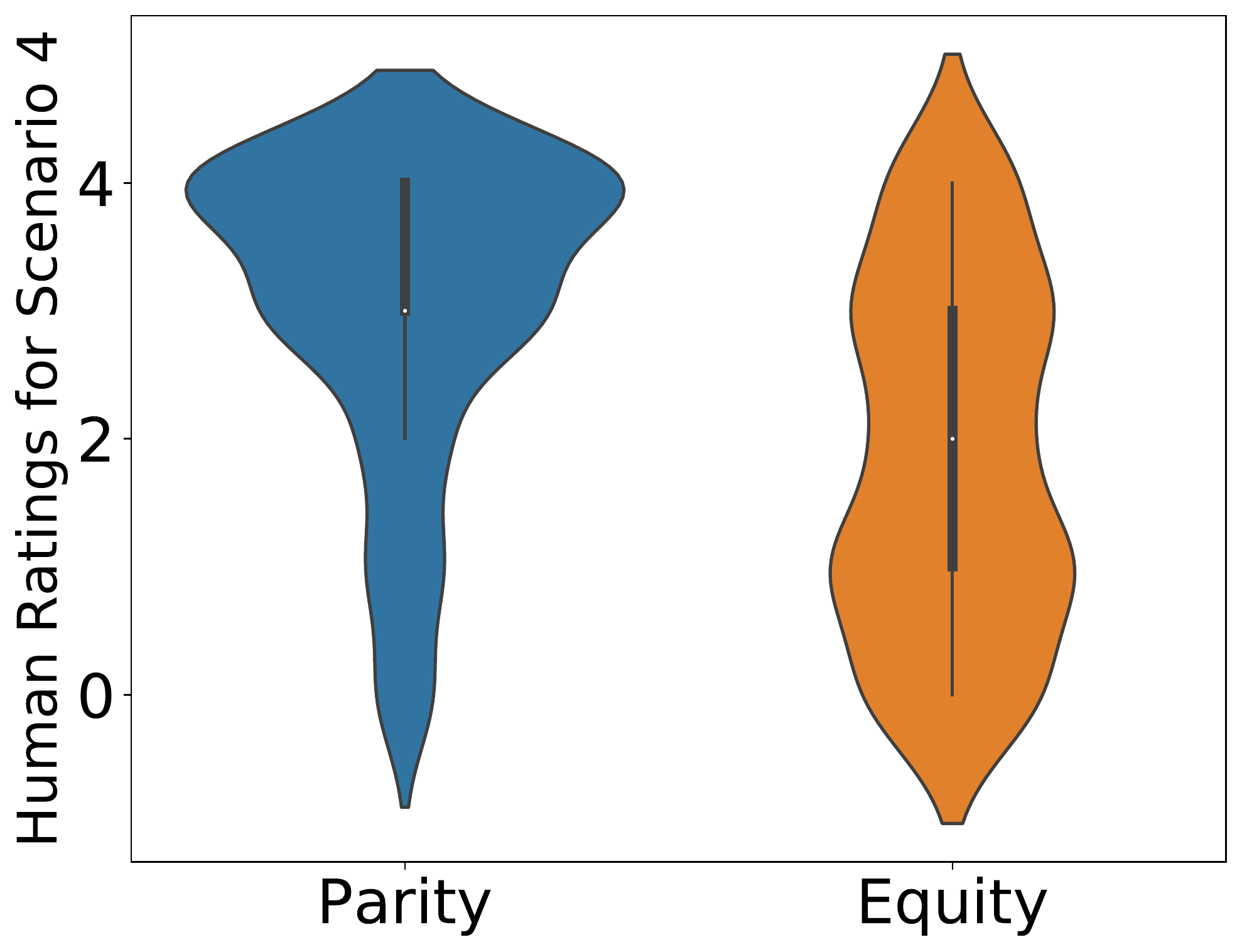}
    \caption{Human ratings of equity and parity notions of fairness in different scenarios.}
    \label{mturk_violin}
\end{figure}

The statistics of ratings for each of the 4 scenarios is shown in Figure~\ref{mturk_violin}. In addition, Table~\ref{votes_stats} depicts the number of mechanical turk workers who preferred a certain solution following a fairness definition in each of the scenarios. Similar to findings in \cite{saxena2019fairness}, we also observed the support for the principle of affirmative action in our experiments which relates to our notion. From the results it is evident that strong preference is given to our notion introduced in this paper for scenarios 1 and 2, and despite the fact that scenarios 3 and 4 are not over-preferred for our notion, there are still considerable number of people who gave preference to our notion in these scenarios. All the justifications written down by the respondents were analyzed. For each preference recorded in this paper, respondents gave justifications that cover a wide range of perspectives. The dataset can be found in \footnotemark[\getrefnumber{projecturl}].

\begin{table}
\begin{tabular}{ p{0.3cm} p{1.4cm} p{1.4cm} p{1.4cm} p{1.4cm}}
 \toprule
& \textbf{Scenario 1} & \textbf{Scenario 2} &\textbf{Scenario 3}&\textbf{Scenario 4}\\
 \midrule
 \parbox[t]{2mm}{\multirow{1}{*}{\rotatebox[origin=c]{90}{Equity}}}
 &\textbf{134} & \textbf{115} &59 &44\\[15pt]
 \midrule
 \parbox[t]{2mm}{\multirow{1}{*}{\rotatebox[origin=c]{90}{Parity}}} &
   16&35&\textbf{91}&\textbf{106}\\[15pt]
 \bottomrule
\end{tabular}
\caption{Number of people preferring solutions provided by the equity vs solutions provided by the parity notions of fairness in different scenarios.}
\label{votes_stats}
\end{table}

\section{Related Work}
With relatively recent popularity of fairness in machine learning and natural language processing domains, the need to find a universal and a more complete fairness definition and measure is crucial. Although finding such definition and measure is a challenge not only in machine learning but also in social and political sciences, steps need to be taken to make current definitions evolve and cover more real world cases. In light of this many fairness definitions have been proposed. Some tried to complement others and some starting a new direction and view-point on their own. Different body of work tried to incorporate the proposed definitions in different downstream tasks such as classification and regression \cite{pmlr-v81-menon18a,
berk2017convex, Krasanakis:2018:ASR:3178876.3186133, agarwal2019fair,goel2018non}.
  
\subsection{Fairness Definitions}
For a more complete list of existing fairness definitions there exists papers that survey \cite{mehrabi2019survey} and explain \cite{verma2018fairness} proposed definitions. Here we will elaborate some important and widely known definitions related to our work introduced in this paper.

\subsubsection{Statistical Parity} In statistical parity \cite{Dwork:2012:FTA:2090236.2090255} the goal is to satisfy $P(\hat{Y} |A = a) = P(\hat{Y}|A = b)$. This notion states that regardless of the belonging demographic group $A$, the predicted outcome should be the same for both demographic groups $A=a$ and $A=b$.
\subsubsection{Equalized Odds} In equalized odds \cite{hardt2016equality}   the goal is to satisfy  $P(\hat{Y}=1|A=a,Y=y) = P(\hat{Y}=1|A=b,Y=y)$ for $y\in\{0,1\}$. This notion states that both groups $A=a$ and $A=b$ should have equal true positive and false positive rates.
\subsubsection{Equal Opportunity} In equal opportunity \cite{hardt2016equality} the goal is to satisfy $P(\hat{Y}=1|A=a,Y=1) = P(\hat{Y}=1|A=b,Y=1)$. This notion states that both demographic groups $A=a$ and $A=b$ should have equal true positive rates.
\subsubsection{Counterfactual Fairness} In counterfactual fairness \cite{NIPS2017_6995} the goal is to satisfy
$P(\hat{Y}_{A\xleftarrow{}a }(U)=y|X=x,A=a) = P(\hat{Y}_{A\xleftarrow{}a'}(U)=y|X =x,A=a)$ under any $X=x$ and $A=a$ for all $y$ and for any value $a'$ feasible for $A$. The perception in counterfactual fairness definition is that if the decision is the same in both the actual and counterfactual world where the individual belonged to a different group then it is called a fair decision. 

\subsection{Fairness Definitions in Classification}
Research in fairness domain does not conclude itself in defining fairness definitions and measures but also incorporation of these definitions in tasks such as classification \cite{calders2010three,huang2019stable}. This incorporation on its own is a challenge. Some methods introduce pre-processing methods that augment the train data for discrimination removal \cite{Kamiran2012}. Some other methods perform in-processing which tries to incorporate the fairness objective during training phase \cite{kamishima2012fairness,zafar2015fairness,wu2018fairnessaware}. Other definitions require post-processing techniques in which discrimination removal is performed after the training phase treating the model as a black box~\cite{hardt2016equality,NIPS2017_7151}. Literature on fair classification also targets a wide variety of fairness definitions, such as equality of opportunity and equalized odds \cite{hardt2016equality,woodworth2017learning}, statistical parity \cite{agarwal2018reductions}, and subgroup fairness \cite{pmlr-v97-ustun19a}, and their incorporation in classification task.
\section{Conclusion}
 In this work, we proposed a definition of fairness based upon equity, demonstrated its appeal as a fairness outcome to a wide audience, and formalized it for classification. We tested this approach in a traditional cross validation setup, and demonstrated how it can be used in a real-world environment, such as unfairness that can arise from the feedback loop. Our results show the effectiveness of our method in mitigating bias and achieving fairness. We also performed human evaluation to evaluate our notion in different scenarios with the equality/parity notion of fairness. As a future direction, our definition can be utilized to achieve and study the effects of equity in classification with different techniques. In this work, we provide a framework for equity to be formalized; however, there is still work to be done in the area of fairness with regards to equity. Future work is to further study how the equity notion interacts with other existing definitions of fairness, such as equality of opportunity, equalized odds or other definitions in the equality domain other than statistical parity. It can also be extended to other machine learning tasks such as regression.

\section{Acknowledgments} 
We wanted to thank Hrayr Harutyunyan and Mozhdeh Gheini for their help and comments.

\bibliographystyle{aaai.bst}
\bibliography{references}
\newpage
\section{Appendixes}
In this section we are going to report some additional and detailed numbers reported in the main paper, such as detailed averaged values shown in Figures \ref{plots1} and \ref{iteration_exp} for the 10 conduced experiments on different splits of data along with the corresponding standard deviations in parenthesis. As also mentioned in the main text, due to the existing variance in different random splits of the dataset, we found reporting the $p$-values with Mann-Whitney U test more suitable; however, here we also report detailed standard deviations for the sake of completeness. We would also include details of our model architecture and also the mechanical turk survey conducted and discussed in the main paper in this section.

\begin{table}[h]
\begin{tabular}{ p{1.4cm} p{5.4cm}}
 \toprule
 Layer Type&Parameters\\
 \midrule
 dense& 256 hidden dimension, tanh activation \\
 dense& 2 output dimension\\
  \bottomrule
\end{tabular}
\caption{Architecture of model used in our experiments.}
    \label{model}
\end{table}

\begin{table}[h]
\begin{tabular}{ c p{0.5cm} p{1.8cm} p{1.8cm} p{1.7cm}}
 \toprule
& \textbf{Beta} & \textbf{Equity} &\textbf{Parity}&\textbf{Classifier}\\
 \midrule
 \parbox[t]{2mm}{\multirow{10}{*}{\rotatebox[origin=c]{90}{Accuracy}}} & 
 0.0&84.76\%(0.41)& 84.76\%(0.41)&84.76\%(0.41)\\[1pt]
   &0.1&84.68\%(0.42)& 84.83\%(0.46)&NA\\[1pt]
   &0.2&84.29\%(0.51)&84.89\%(0.45) &NA\\[1pt]
   &0.3&83.51\%(0.45)& 84.73\%(0.50)&NA\\[1pt]
   &0.4&82.86\%(0.45)&84.55\%(0.51) &NA\\[1pt]
 & 0.5&82.00\%(0.49)& 84.36\%(0.57)&NA\\[1pt]
 &0.6&81.48\%(0.43)& 84.14\%(0.49)&NA\\[1pt]
 &0.7&80.81\%(0.47)&83.97\%(0.57) &NA\\[1pt]
 &0.8&80.45\%(0.63)& 83.74\%(0.44)&NA\\[1pt]
  &0.9&79.38\%(1.12)& 83.71\%(0.54)&NA\\[1pt]
 \bottomrule
 \parbox[t]{2mm}{\multirow{10}{*}{\rotatebox[origin=c]{90}{Fairness Gain}}} & 
   0.0&0.00\%(0.00)& 0.00\%(0.00)&0.00\%(0.00)\\[1pt]
   &0.1&0.61\%(0.06)& 0.30\%(0.04)&NA\\[1pt]
   &0.2&1.24\%(0.10)& 0.58\%(0.06)&NA\\[1pt]
   &0.3&1.80\%(0.16)& 0.84\%(0.07)&NA\\[1pt]
   &0.4&2.25\%(0.16)& 1.08\%(0.09)&NA\\[1pt]
 & 0.5&2.61\%(0.17)& 1.30\%(0.13)&NA\\[1pt]
 &0.6&2.83\%(0.21)& 1.42\%(0.17)&NA\\[1pt]
 &0.7&3.12\%(0.24)& 1.58\%(0.11)&NA\\[1pt]
 &0.8&3.33\%(0.28)& 1.67\%(0.13)&NA\\[1pt]
  &0.9&3.77\%(0.42)&1.86\%(0.12) &NA\\[1pt]
  \bottomrule
\end{tabular}
\caption{Averaged percent accuracy and fairness gain for the Adult dataset along with the standard deviation numbers reported in parenthesis for different $\beta$ values.}
\label{Adult_acc_gain}
\end{table}

\begin{table}[h]
\begin{tabular}{c p{0.5cm} p{1.8cm} p{1.8cm} p{1.7cm}}
 \toprule
& \textbf{Beta} & \textbf{Equity} &\textbf{Parity}&\textbf{Classifier}\\
 \midrule
 \parbox[t]{2mm}{\multirow{10}{*}{\rotatebox[origin=c]{90}{Accuracy}}} & 
 0.0&66.80\%(2.19)& 66.80\%(2.19)&66.80\%(2.19)\\[2pt]
   &0.1&66.60\%(1.59)& 67.04\%(1.90)&NA\\[1pt]
   &0.2&66.23\%(1.78)&66.47\%(2.08) &NA\\[1pt]
   &0.3&65.66\%(2.04)& 66.48\%(1.87)&NA\\[1pt]
   &0.4&65.39\%(1.82)&66.30\%(1.74) &NA\\[1pt]
 & 0.5&65.17\%(1.84)&66.41\%(1.22)&NA\\[1pt]
 &0.6&64.61\%(2.12)&66.51\%(1.43)&NA\\[1pt]
 &0.7&64.25\%(2.37)&66.51\%(1.49) &NA\\[1pt]
 &0.8&64.10\%(2.08)&66.61\%(1.49) &NA\\[1pt]
  &0.9&64.06\%(1.62)&66.39\%(1.17)&NA\\[1pt]
 \bottomrule
 \parbox[t]{2mm}{\multirow{10}{*}{\rotatebox[origin=c]{90}{Fairness Gain}}} & 
   0.0&0.0\%(0.00)& 0.0\%(0.00)&0.0\%(0.00)\\[1pt]
   &0.1&1.80\%(0.52)& 0.77\%(0.32)&NA\\[1pt]
   &0.2&3.26\%(0.57)& 1.38\%(0.53)&NA\\[1pt]
   &0.3&4.07\%(0.61)& 1.72\%(0.44)&NA\\[1pt]
   &0.4&4.48\%(0.48)& 1.94\%(0.43)&NA\\[1pt]
 & 0.5&4.81\%(0.65)& 2.26\%(0.59)&NA\\[1pt]
 &0.6&5.11\%(0.72)& 2.53\%(0.50)&NA\\[1pt]
 &0.7&5.38\%(0.80)& 2.07\%(0.38)&NA\\[1pt]
 &0.8&5.37\%(1.06)& 2.78\%(0.70)&NA\\[1pt]
  &0.9&6.05\%(1.40)& 2.91\%(0.36)&NA\\[1pt]
  \bottomrule
\end{tabular}
\caption{Averaged percent accuracy and fairness gain for the COMPAS dataset along with the standard deviation numbers reported in parenthesis for different $\beta$ values.}
\label{compas_acc_gain}
\end{table}

\begin{table}[h]
\begin{tabular}{|p{0.5cm} |c||p{1.07cm}|p{1.15cm}||p{1.07cm}|p{1.15cm}|  }
 \hline
 \multicolumn{2}{|c||}{}&\multicolumn{2}{c||}{COMPAS Dataset}&\multicolumn{2}{c|}{Adult Dataset}\\
 \hline
 \multicolumn{2}{|c||}{}&\multicolumn{2}{c||}{$p$-value}&\multicolumn{2}{c|}{$p$-value}\\
 \hline
Iter& & Parity &Classifier&Parity&Classifier\\
 \hline
1&\multirow{8}{*}{\rotatebox[origin=c]{90}{Equity}}   & 0.3387    &0.2854  & 0.3115&0.2603 \\
 2&    & 0.2248 &0.2137 & 0.2137& 0.0520\\
3&&0.1365 &0.1207  &0.1365&0.0106\\
4&&0.1207 & 0.0320 &0.1061&0.0011\\
5&&0.0929 & 0.0070 &0.0445&0.0018\\
6& & 0.0269&  0.0008&0.0226&0.0006\\
7& & 0.0378&0.0004  &0.0445&0.0004\\
8&& 0.0106& 0.0004 &0.0226&0.0004\\
9&&0.0106 & 0.0004 &0.0106&0.0001\\
 \hline
\end{tabular}
\caption{Performance of Mann-Whitney U test for showing the effectiveness of Equity in reducing bias in the feedback loop compared to Parity and Classifier losses over different iterations for COMPAS and Adult datasets. The obtained $p$-values show the significance of our reported results in Figure \ref{iteration_exp} for $\beta$ value of 0.1.}
\label{01_feedbackloop_p}
\end{table}

\begin{table}[h]
\begin{tabular}{|p{0.5cm} |c||p{1.07cm}|p{1.15cm}||p{1.07cm}|p{1.15cm}|  }
 \hline
 \multicolumn{2}{|c||}{}&\multicolumn{2}{c||}{COMPAS Dataset}&\multicolumn{2}{c|}{Adult Dataset}\\
 \hline
 \multicolumn{2}{|c||}{}&\multicolumn{2}{c||}{$p$-value}&\multicolumn{2}{c|}{$p$-value}\\
 \hline
Iter& & Parity &Classifier&Parity&Classifier\\
 \hline
1&\multirow{8}{*}{\rotatebox[origin=c]{90}{Equity}}   & 0.1537    & 0.0606 & 0.0156&0.0004 \\
2&    & 0.0606 & 0.0056& 0.0005&9.1e-05 \\
3&&0.0156 & 0.0003 &0.0001&9.1e-05\\
4&&0.0045 & 0.0001 &9.1e-05&9.1e-05\\
5&&0.0018 & 9.1e-05 &9.1e-05&9.1e-05\\
6& & 0.0005& 9.1e-05 &9.1e-05&9.1e-05\\
7& &0.0005 & 9.1e-05 &9.1e-05&9.1e-05\\
8&&0.0002 & 9.1e-05 &9.1e-05&9.1e-05\\
9&&0.0002 & 9.1e-05 &9.1e-05&9.1e-05\\
 \hline
\end{tabular}
\caption{Performance of Mann-Whitney U test for showing the effectiveness of Equity in reducing bias in the feedback loop compared to Parity and Classifier losses over different iterations for COMPAS and Adult datasets. The obtained $p$-values show the significance of our reported results in Figure \ref{iteration_exp} for $\beta$ value of 0.9.}
\label{09_feedbackloop_p}
\end{table}

\begin{table}[h]
\begin{tabular}{ c p{0.5cm} p{1.7cm} p{1.7cm} p{1.7cm}}
 \toprule
& \textbf{Beta} & \textbf{Equity} &\textbf{Parity}&\textbf{Classifier}\\
 \midrule
 \parbox[t]{2mm}{\multirow{10}{*}{\rotatebox[origin=c]{90}{Bias}}} & 
 0&11.82\%(0.78)&11.82\%(0.78) &11.82\%(0.78)\\[1pt]
   &1&11.79\%(0.74)& 11.87\%(0.77)&11.93\%(0.79)\\[1pt]
   &2&11.80\%(0.74)& 12.04\%(0.79)&12.10\%(0.76)\\[1pt]
   &3&11.82\%(0.75)& 12.18\%(0.80)&12.30\%(0.72)\\[1pt]
   &4&11.79\%(0.69)&12.25\%(0.76) &12.47\%(0.67)\\[1pt]
 & 5&11.84\%(0.69)&12.40\%(0.77) &12.69\%(0.69)\\[1pt]
 &6&11.70\%(0.60)&12.39\%(0.73) &12.77\%(0.64)\\[1pt]
 &7&11.67\%(0.70)&12.47\%(0.85) &12.98\%(0.75)\\[1pt]
 &8&11.67\%(0.67)&12.55\%(0.82) &13.16\%(0.70)\\[1pt]
  &9&11.66\%(0.63)&12.59\%(0.81) &13.24\%(0.70)\\[1pt]
 \bottomrule
\end{tabular}
\caption{Detailed averaged percent biases and standard deviation results in parenthesis for the COMPAS dataset shown in Figure \ref{iteration_exp} for $\beta$ value of 0.1.}
\label{compas_feed_0.1}
\end{table}

\begin{table}[h]
\begin{tabular}{ c p{0.5cm} p{1.7cm} p{1.7cm} p{1.7cm}}
 \toprule
& \textbf{Beta} & \textbf{Equity} &\textbf{Parity}&\textbf{Classifier}\\
 \midrule
 \parbox[t]{2mm}{\multirow{10}{*}{\rotatebox[origin=c]{90}{Bias}}} & 
 0&11.18\%(0.78)& 11.82\%(0.78)&11.82\%(0.78)\\[1pt]
   &1&11.45\%(0.73)& 11.75\%(0.73)&11.93\%(0.79)\\[1pt]
   &2&11.18\%(0.70)& 11.77\%(0.77)&12.10\%(0.76)\\[1pt]
   &3&10.90\%(0.69)& 11.74\%(0.77)&12.30\%(0.72)\\[1pt]
   &4&10.63\%(0.75)& 11.68\%(0.77)&12.47\%(0.67)\\[1pt]
 & 5&10.40\%(0.76)& 11.71\%(0.75)&12.69\%(0.69)\\[1pt]
 &6&10.11\%(0.73)&11.52\%(0.64) &12.77\%(0.64)\\[1pt]
 &7&9.85\%(0.81)&11.46\%(0.77) &12.98\%(0.75)\\[1pt]
 &8&9.66\%(0.75)&11.40\%(0.77) &13.16\%(0.70)\\[1pt]
  &9&9.35\%(0.78)&11.36\%(0.72) &13.24\%(0.70)\\[1pt]
 \bottomrule
\end{tabular}
\caption{Detailed averaged percent biases and standard deviation results in parenthesis for the COMPAS dataset shown in Figure \ref{iteration_exp} for $\beta$ value of 0.5.}
\label{compas_feed_0.5}
\end{table}

\begin{table}[h]
\begin{tabular}{  c p{0.5cm} p{1.7cm} p{1.7cm} p{1.7cm}}
 \toprule
& \textbf{Beta} & \textbf{Equity} &\textbf{Parity}&\textbf{Classifier}\\
 \midrule
 \parbox[t]{2mm}{\multirow{10}{*}{\rotatebox[origin=c]{90}{Bias}}} & 
 0&11.82\%(0.78)& 11.82\%(0.78)&11.82\%(0.78)\\[1pt]
   &1&11.38\%(0.86)& 11.69\%(0.75)&11.93\%(0.79)\\[1pt]
   &2&10.97\%(0.84)& 11.64\%(0.76)&12.10\%(0.76)\\[1pt]
   &3&10.59\%(0.91)& 11.57\%(0.75)&12.30\%(0.72)\\[1pt]
   &4&10.20\%(1.03)&11.48\%(0.71) &12.47\%(0.67)\\[1pt]
 & 5&9.92\%(1.04)& 11.41\%(0.73)&12.69\%(0.69)\\[1pt]
 &6&9.53\%(1.08)& 11.21\%(0.63)&12.77\%(0.64)\\[1pt]
 &7&9.19\%(1.18)& 11.12\%(0.74)&12.98\%(0.75)\\[1pt]
 &8&8.91\%(1.16)&11.06\%(0.71) &13.16\%(0.70)\\[1pt]
  &9&8.52\%(1.21)&11.01\%(0.74) &13.24\%(0.70)\\[1pt]
 \bottomrule
\end{tabular}
\caption{Detailed averaged percent biases and standard deviation results in parenthesis for the COMPAS dataset shown in Figure \ref{iteration_exp} for $\beta$ value of 0.9.}
\label{compas_feed_0.9}
\end{table}

\begin{table}[h]
\begin{tabular}{  c p{0.5cm} p{1.7cm} p{1.7cm} p{1.7cm}}
 \toprule
& \textbf{Beta} & \textbf{Equity} &\textbf{Parity}&\textbf{Classifier}\\
 \midrule
 \parbox[t]{2mm}{\multirow{10}{*}{\rotatebox[origin=c]{90}{Bias}}} & 
 0&19.91\%(0.18)&19.91\%(0.18) &19.91\%(0.18)\\[1pt]
   &1&19.83\%(0.17)&19.86\%(0.17) &19.89\%(0.16)\\[1pt]
   &2&19.75\%(0.17)& 19.81\%(0.16)&19.88\%(0.16)\\[1pt]
   &3&19.67\%(0.17)& 19.76\%(0.17)&19.87\%(0.17)\\[1pt]
   &4&19.59\%(0.16)&19.70\%(0.16) &19.86\%(0.16)\\[1pt]
 & 5&19.51\%(0.19)&19.67\%(0.18) &19.85\%(0.18)\\[1pt]
 &6&19.43\%(0.20)&19.63\%(0.19) &19.83\%(0.19)\\[1pt]
 &7&19.34\%(0.22)&19.58\%(0.21) &19.81\%(0.20)\\[1pt]
 &8&19.26\%(0.23)&19.53\%(0.21) &19.79\%(0.21)\\[1pt]
  &9&19.18\%(0.23)&19.48\%(0.21)&19.77\%(0.21)\\[1pt]
 \bottomrule
\end{tabular}
\caption{Detailed averaged percent biases and standard deviation results in parenthesis for the Adult dataset shown in Figure \ref{iteration_exp} for $\beta$ value of 0.1.}
\label{adult_feed_0.1}
\end{table}

\begin{table}[h]
\begin{tabular}{ c p{0.5cm} p{1.7cm} p{1.7cm} p{1.7cm}}
 \toprule
& \textbf{Beta} & \textbf{Equity} &\textbf{Parity}&\textbf{Classifier}\\
 \midrule
 \parbox[t]{2mm}{\multirow{10}{*}{\rotatebox[origin=c]{90}{Bias}}} & 
 0&19.91\%(0.18)& 19.91\%(0.18)&19.91\%(0.18)\\[1pt]
   &1&19.61\%(0.16)& 19.76\%(0.17)&19.89\%(0.16)\\[1pt]
   &2&19.32\%(0.18)& 19.59\%(0.17)&19.88\%(0.16)\\[1pt]
   &3&19.01\%(0.17)&19.42\%(0.17) &19.87\%(0.17)\\[1pt]
   &4&18.73\%(0.15)&19.30\%(0.16) &19.86\%(0.16)\\[1pt]
 & 5&18.44\%(0.16)&19.12\%(0.17) &19.85\%(0.18)\\[1pt]
 &6&18.71\%(0.16)& 18.97\%(0.16)&19.83\%(0.19)\\[1pt]
 &7&17.90\%(0.18)& 18.82\%(0.18)&19.81\%(0.20)\\[1pt]
 &8&17.63\%(0.20)& 18.66\%(0.18)&19.79\%(0.21)\\[1pt]
  &9&17.38\%(0.21)&18.51\%(0.19) &19.77\%(0.21)\\[1pt]
 \bottomrule
\end{tabular}
\caption{Detailed averaged percent biases and standard deviation results in parenthesis for the Adult dataset shown in Figure \ref{iteration_exp} for $\beta$ value of 0.5.}
\label{adult_feed_0.5}
\end{table}

\begin{table}[h]
\begin{tabular}{ c p{0.5cm} p{1.7cm} p{1.7cm} p{1.7cm}}
 \toprule
& \textbf{Beta} & \textbf{Equity} &\textbf{Parity}&\textbf{Classifier}\\
 \midrule
 \parbox[t]{2mm}{\multirow{10}{*}{\rotatebox[origin=c]{90}{Bias}}} & 
 0&19.91\%(0.18)&19.91\%(0.18) &19.91\%(0.18)\\[1pt]
   &1&19.48\%(0.21)& 19.70\%(0.17)&19.89\%(0.16)\\[1pt]
   &2&19.08\%(0.26)&19.47\%(0.17) &19.88\%(0.16)\\[1pt]
   &3&18.64\%(0.28)&19.24\%(0.17) &19.87\%(0.17)\\[1pt]
   &4&18.25\%(0.29)& 19.04\%(0.16)&19.86\%(0.16)\\[1pt]
 & 5&17.85\%(0.29)&18.83\%(0.17) &19.85\%(0.18)\\[1pt]
 &6&17.46\%(0.31)& 18.63\%(0.17)&19.83\%(0.19)\\[1pt]
 &7&17.09\%(0.34)& 18.43\%(0.18)&19.81\%(0.20)\\[1pt]
 &8&16.70\%(0.37)&18.22\%(0.19) &19.79\%(0.21)\\[1pt]
  &9&16.33\%(0.40)&18.03\%(0.21) &19.77\%(0.21)\\[1pt]
 \bottomrule
\end{tabular}
\caption{Detailed averaged percent biases and standard deviation results in parenthesis for the Adult dataset shown in Figure \ref{iteration_exp} for $\beta$ value of 0.9.}
\label{adult_feed_0.9}
\end{table}

\begin{figure}[h]
\centering
\includegraphics[width=0.5\textwidth,trim=26cm 33cm 26cm 0cm,clip=true]{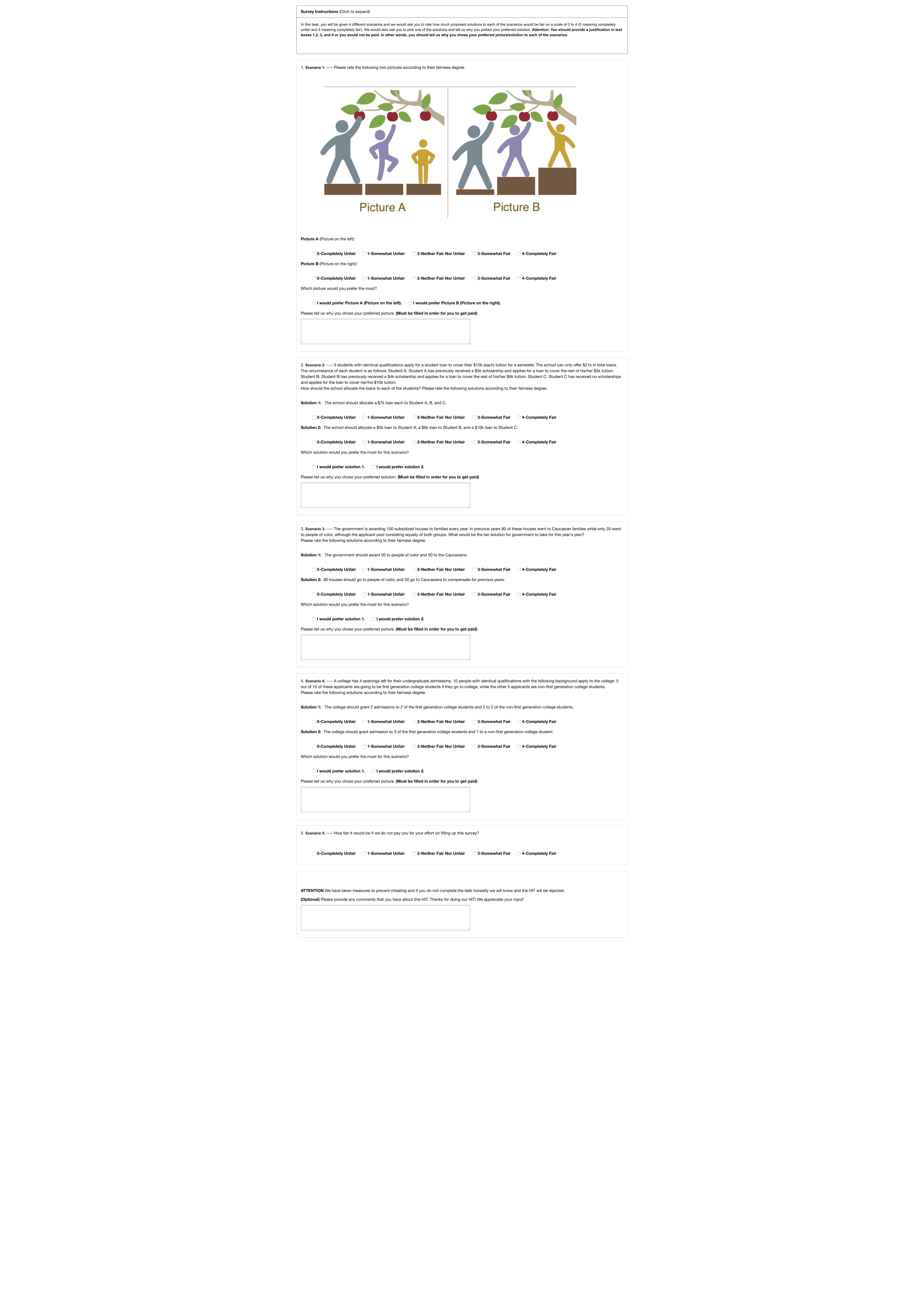}
\caption{Survey questionnaire used in the mechanical turk experiment.}
\label{survey}
\end{figure}

\end{document}